\documentclass[sigconf, nonacm]{acmart}

\usepackage{algorithm}
\usepackage{algpseudocode}
\usepackage{listings}
\usepackage{float}
\usepackage{comment}
\usepackage{placeins}
\usepackage{hyperref}

\AtBeginDocument{%
  }

\copyrightyear{2026}
\acmYear{2026}
\setcopyright{rightsretained}
\acmConference[FDG '26]{International Conference on the Foundations of Digital Games}{August 10--13, 2026}{Copenhagen, Denmark}
\acmBooktitle{International Conference on the Foundations of Digital Games (FDG '26), August 10--13, 2026, Copenhagen, Denmark}
\acmPrice{}
\acmDOI{XXXXXXX.XXXXXXX}
\acmISBN{/26/08}

\begin{document}

\title{How Far Can We Go with Pixels Alone? A Pilot Study on Screen-Only Navigation in Commercial 3D ARPGs}

\author{Kaijie Xu}
\affiliation{%
  \institution{McGill University}
  \city{Montreal}
  \state{Quebec}
  \country{Canada}}
\email{kaijie.xu2@mail.mcgill.ca}

\author{Mustafa Bugti}
\affiliation{%
  \institution{Connecticut College}
  \city{New London}
  \state{Connecticut}
  \country{USA}}
\email{mbugti@conncoll.edu}

\author{Clark Verbrugge}
\affiliation{%
  \institution{McGill University}
  \city{Montreal}
  \state{Quebec}
  \country{Canada}}
\email{clump@cs.mcgill.ca}

\begin{abstract}
Modern 3D game levels rely heavily on visual guidance, yet the navigability of level layouts remains difficult to quantify. Prior work either simulates play in simplified environments or analyzes static screenshots for visual affordances, but neither setting faithfully captures how players explore complex, real-world game levels. In this paper, we build on an existing open-source visual affordance detector and instantiate a screen-only exploration and navigation agent that operates purely from visual affordances. Our agent consumes live game frames, identifies salient interest points, and drives a simple finite-state controller over a minimal action space to explore Dark Souls–style linear levels and attempt to reach expected goal regions. Pilot experiments show that the agent can traverse most required segments and exhibits meaningful visual navigation behavior, but also highlight that limitations of the underlying visual model prevent truly comprehensive and reliable auto-navigation. We argue that this system provides a concrete, shared baseline and evaluation protocol for visual navigation in complex games, and we call for more attention to this necessary task. Our results suggest that purely vision-based sense-making models, with discrete single-modality inputs and without explicit reasoning, can effectively support navigation and environment understanding in idealized settings, but are unlikely to be a general solution on their own.
\end{abstract}



\maketitle

\section{Introduction}

Modern 3D games rely heavily on visual guidance to communicate where players can and should go. In realistic, visually dense levels, designers shape routes through landmarks, lighting, composition, and subtle contrast rather than explicit arrows or waypoints \cite{guo2023visual, milam2010analysis, hullett2012science}. Souls-like action RPGs are a particularly demanding case: they often minimize HUD markers while using environmental silhouettes, skyline composition, and path textures to pull players toward boss arenas or critical shortcuts \cite{farraj_between_2023}. Despite this central role of visual guidance in level design, it remains difficult to operationalize and measure navigability in real, shipped games. Most existing navigation agents are trained in simulators such as AI2-THOR, Gibson, or Habitat \cite{kolve2017ai2, xia2018gibson, savva2019habitat}, or in stylized research platforms like ViZDoom, Project Malmo (Minecraft), or similar 3D benchmarks \cite{kempka2016vizdoom, johnson2016malmo, mirowski2016learning, pathak2017curiosity, lample2017playing, baker2022video}. These systems validate that agents can navigate complex 3D environments from pixels using deep reinforcement learning, mapping, and memory, but they typically assume full access to simulator state, structured tasks such as PointGoal or ObjectNav \cite{anderson2018evaluation}, and clean sensor channels that differ significantly from the messy, overdrawn visuals of commercial games. In parallel, commercial automated playtesting tools focus on coverage and pacing with engine access rather than screen-only sensing \cite{prasetya2020navigation}.

Within games research, a complementary line of work examines visual guidance qualitatively, studying how composition and affordances steer human players through open worlds and FPS levels \cite{guo2023visual, milam2010analysis, hullett2012science, farraj_between_2023}. More recently, a Visual Navigation Assistant for 3D RPGs \cite{xu2025adaptive} detects visually defined spatial transition points (STPs) from single frames and selects a main spatial transition point (MSTP) to guide human players. In their terminology, an STP is any traversable doorway, ladder, corridor, or passage that links two distinct map regions, while an MSTP is the unique STP on the designer-defined critical path toward the current macro-objective (typically the next boss gate or primary level goal). Their two-stage detector-selector pipeline, trained on annotated screenshots from multiple ARPGs, shows that modern levels contain stable visual affordances that can be learned and overlaid as in-game hints. However, that work remains strictly human-in-the-loop: the model runs on isolated frames, does not reason over temporal context or failures, and is only evaluated as an on-screen guidance overlay rather than as the perception stack of an autonomous agent.

In this paper we take a step further and ask whether a game can be navigated \emph{purely} from visual affordances. We treat a 3D RPG game as an embodied navigation problem with a minimal, discrete action space and no access to game coordinates, meshes, or engine APIs. Building directly on the existing STP/MSTP pipeline as a fixed perception backbone, we design a finite-state controller with progress signals and a lightweight visual memory bank that tries to turn single-frame affordance predictions into screen-only navigation behavior. To evaluate this agent in real levels, we introduce a milestone-based protocol that defines a designer-specified route as a sequence of recognizable viewpoints and uses image matching against these milestones to segment and score runs without any internal game state. We instantiate this setup in real game environments, treating them as a testbed where visual guidance is both critical to human play and challenging for automated systems.

Our pilot experiments suggest that this approach is promising but fundamentally constrained. The agent can traverse many required segments and exhibits meaningful visual navigation behavior under favorable conditions, but it also fails in visually ambiguous or extreme viewpoints where the underlying STP/MSTP model struggles. These results indicate that purely vision-based sense-making models, under discrete single-modality input and without explicit reasoning, can effectively support navigation and environment understanding in idealized segments, yet are unlikely to constitute a general solution on their own. At the same time, our agent and evaluation protocol \footnote{\url{https://github.com/Nortrom1213/VisualNavigation}} provide a concrete, reusable baseline for \emph{screen-only} visual navigation in modern games, and we hope they encourage more work at the intersection of visual affordances, level design, and embodied agents. Our contributions are:
\begin{itemize}
  \item We build on prior visual affordance models to construct a general, screen-only navigation controller that operates in modern 3D ARPG levels using only STP/MSTP predictions and discrete camera/forward actions.
  \item We propose a practical milestone-based evaluation protocol and toolchain that defines navigation tasks as ordered sequences of visual checkpoints and scores agents purely from image matching, without engine instrumentation.
  \item We conduct a pilot study on multiple Souls-like ARPG levels that validate both the feasibility of fully visual navigation in many segments and the limitations of current single-frame affordance models, supported by quantitative metrics and qualitative case studies.
\end{itemize}
\section{Related Works}

\subsection{Navigation Agent}

A large body of work studies agents that navigate 3D environments from visual inputs alone with a discrete action space. Zhu et al.\ introduce target-driven visual navigation, where an agent receives the current view and an image of a goal and learns via deep reinforcement learning to reach the goal in AI2-THOR scenes \cite{zhu2017target, kolve2017ai2}. Mirowski et al.\ treat navigation in complex 3D mazes as an RL problem \cite{mirowski2016learning}, while Pathak et al.\ use curiosity-based rewards so that agents in VizDoom and Super Mario explore states where their feature predictions are wrong \cite{pathak2017curiosity}. These works rely on simulators for embodied AI such as AI2-THOR, Gibson, and Habitat, which offer photorealistic indoor scenes and standardized tasks and metrics \cite{kolve2017ai2, xia2018gibson, savva2019habitat, anderson2018evaluation}. Later work adds mapping and memory. Cognitive Mapping and Planning (CMP) learns a top-down map from first-person images and runs a differentiable planner on this map \cite{gupta2017cognitive}. Active Neural SLAM combines a learned SLAM module with a classical planner and hierarchical policies for exploration from RGB input \cite{chaplot2020learning}, and Neural Topological SLAM instead learns a topological graph of visited locations for image-goal navigation \cite{chaplot2020neural}. Ramakrishnan et al.\ predict free space beyond the current field of view and use this occupancy anticipation to guide navigation \cite{ramakrishnan2020occupancy}.

Evaluation practice has also become more standardized. Anderson et al.\ define embodied navigation tasks (PointGoal, ObjectGoal, etc.) and metrics such as Success weighted by Path Length (SPL), emphasizing generalization to unseen environments \cite{anderson2018evaluation}. Habitat adopts these definitions and shows that, with enough experience, learned agents can outperform classical SLAM baselines on PointGoal navigation while still using only sensor observations \cite{savva2019habitat, anderson2018evaluation}. Work on hard-exploration problems complements this by decoupling exploration from exploitation: Go-Explore keeps an archive of promising cells, reliably returns to a chosen cell, and then explores outward to discover new behaviors in sparse-reward Atari games \cite{ecoffet2019go, ecoffet2021first}. Variants such as Latent Go-Explore and latent-map exploration use learned state abstractions and reset mechanisms to further enlarge coverage in long-horizon settings \cite{gallouedec2023cell, shah2022rapid}, similar in spirit to our use of milestone saves to restart from designer-defined visual checkpoints. Together, these works motivate our view of Souls-like games as embodied navigation problems driven only by visual input and a small set of camera/forward actions.

\subsection{Visual Research in Games}

Video games have long served as testbeds for vision-based navigation and exploration. In ViZDoom, agents learn to move, shoot, and collect items directly from RGB frames using deep reinforcement learning \cite{kempka2016vizdoom}. Lample and Chaplot extend this with an actor-critic model and curriculum learning to handle harder maze-like maps \cite{lample2017playing}, and curiosity-driven methods further encourage exploration in sparse-reward settings \cite{pathak2017curiosity}. Outside Doom-like games, Malmo exposes Minecraft through a first-person camera and low-level controls, allowing RL agents to explore large voxel worlds \cite{johnson2016malmo}. More recently, Video Pre-Training shows that large models trained on human gameplay videos can map pixels and keypress histories to plausible action sequences in Minecraft, imitating long-horizon navigation from raw video \cite{baker2022video}. CARMI instead drives scripted agents through levels to measure coverage and pacing, but typically relies on engine-level data rather than purely screen-based sensing like SIMA 2, which uses large multimodal foundation models to understand and act in a 3D virtual world \cite{prasetya2020navigation, bolton2025sima}.

A parallel line of work analyzes how level visuals guide human players. Guo systematizes how landmarks, lighting, composition, and color contrast steer players along intended routes in open worlds \cite{guo2023visual}. Milam and Seif El-Nasr’s ``push \& pull'' framework surveys 21 commercial games to show how recurring level-design patterns attract or repel player attention and shape pacing \cite{milam2010analysis}. Hullett links geometry, affordances, and item placement to observed trajectories in FPS levels, giving empirical evidence for layout-behavior effects \cite{hullett2012science}. A focused study of \emph{Elden Ring} shows how subtle cues, such as silhouettes, skyline composition and path textures, can replace explicit markers to guide exploration, but also make navigation demanding and opaque \cite{farraj_between_2023}.

Recent work has started to quantify and operationalize visual guidance cues for automated tools. Xu et al. detect visually-defined spatial transition points (STPs) from single frames in Souls-like games, showing that a two-stage detector-selector pipeline trained on annotated screenshots can correctly identify designer-intended exits and route continuations across multiple ARPGs \cite{xu2025adaptive}. Their pilot study uses STP predictions as an in-game guidance overlay for human players, achieving a high success rate on real levels while also surfacing failure cases where misleading affordances confuse the model. In contrast, existing playtesting systems such as CARMI focus on coverage and pacing with engine-level access \cite{prasetya2020navigation}, and traditional level-design analyses remain human-centered \cite{guo2023visual, milam2010analysis, hullett2012science}. Our project extends these directions by treating visual guidance itself as the sole sensor modality: instead of merely suggesting paths to a human, we build a fully automatic agent that reads only the screen, presses a restricted set of keys, and must still navigate levels from start positions to goal areas. By combining the existing visual pipeline with memory over past visual failures and a milestone-based evaluation tied to recognizable viewpoints, we position our agent as a screen-only visual navigation baseline that both connects to prior work on visual affordances and stress-tests how far purely perceptual cues can carry autonomous exploration in complex, visually dense 3D game worlds. 

\section{Methods}

Our agent is built on top of an existing visual guidance model \cite{xu2025adaptive} that detects \emph{spatial transition points} (STPs) from single frames of gameplay and selects the \emph{main spatial transition point} (MSTP). We keep this perception stack fixed and focus our contributions on the control and memory layers that turn MSTP predictions into screen-only navigation behavior. Concretely, at each frame we select an MSTP target using the fixed perception stack and then apply our control to convert that target into camera and movement actions using only on-screen observations.

\subsection{Base Visual Model: STP/MSTP Perception}

We adopt a two-stage visual model for detecting candidate exits, turns, and
corridor continuations from a single RGB frame $I_t$:

\textbf{(1) STP detector:} A region-based detector (Faster R-CNN with lightweight adapters) takes $I_t$ and predicts $N_t$ candidate
    STPs:
    \[
        \mathcal{C}_t = \left\{ \bigl(b_i, s^{\text{det}}_i, z_i\bigr)
        \right\}_{i=1}^{N_t},
    \]
    where $b_i = (x^i_1, y^i_1, x^i_2, y^i_2)$ is a bounding box in screen
    coordinates, $s^{\text{det}}_i$ is a detector confidence, and $z_i \in
    \mathbb{R}^d$ is a feature embedding extracted from the cropped
    region (via a ResNet-based trunk).
    
\textbf{(2) MSTP selector:} A selector consumes both local and
    global context. Let $g_t \in \mathbb{R}^D$ be a global feature of the full
    frame (e.g., a CNN embedding). For each candidate $i$, the selector
    predicts a scalar
    \[
        s^{\text{sel}}_i = f_\theta\bigl(z_i, g_t, \phi(b_i), h^{\text{hist}}_t
        \bigr),
    \]
    where $\phi(b_i)$ encodes geometric properties (center, width, height,
    sector index), and $h^{\text{hist}}_t$ summarizes recent STP statistics
    (e.g., a histogram of sectors over the last few frames). In our
    implementation, the learned selector network operates on $(z_i, g_t)$,
    while $\phi(b_i)$ and $h^{\text{hist}}_t$ are instantiated as additional
    hand-crafted scoring terms (e.g., sector priors and free-space measures)
    added to the selector score downstream.
    The overall score for candidate $i$ is
    \[
        s_i = \alpha s^{\text{sel}}_i
            + \beta s^{\text{ret}}_i
            + \gamma s^{\text{temp}}_i
            + \psi\bigl(\phi(b_i), h^{\text{hist}}_t\bigr),
    \]
    where $s^{\text{ret}}_i$ is an optional retrieval-based score against an
    offline STP database (if available), $s^{\text{temp}}_i$ encodes
    temporal consistency with the previous MSTP (e.g., via IoU and sector
    priors), and $\psi(\cdot)$ collects geometric and free-space heuristics
    (such as base alignment with walkable regions). Since MSTP is chosen by 
    $\arg\max_i s_i$, $\alpha,\beta,\gamma$ need not be normalized, and $\psi$ 
    is an unnormalized additive heuristic used for re-ranking.
    The MSTP index is
    $
        j^\star = \arg\max_{i} s_i,
        \text{MSTP}_t = b_{j^\star}.
    $

Intuitively, STPs are candidate ``where to go next'' locations visible in the
current frame (doorways, ramps, openings), and the MSTP is the single most
promising continuation. Our navigation agent treats the MSTP center
$\mathbf{c}_t \in \mathbb{R}^2$ as the current target on screen and \emph{only}
observes $(I_t, \text{MSTP}_t)$ at each time-step; it never receives access to
game coordinates, collision meshes, or any other internal state.
We discretize the horizontal field of view into $K$ sectors
(e.g., $K=8$). Each candidate box $b_i$ is assigned a sector index
$\sigma(b_i) \in \{1,\dots,K\}$ by its center $x$-coordinate. This sector label
is used by both the controller (for steering) and, when enabled, the memory
bank (for scoring past decisions by direction).

\subsection{Screen-Only Controller}

Our controller turns $(I_t, \text{MSTP}_t)$ into a sequence of discrete camera and movement actions $\mathcal{A} = \{\texttt{U},\texttt{I},\texttt{O},\texttt{P}, \texttt{W},\varnothing\}$. Here $\texttt{U},\texttt{I},\texttt{O},\texttt{P}$ denote short camera pulses (left / up / right / down) and $\texttt{W}$ toggles forward movement. In practice these pulses are bound to otherwise unused keys that we remap to camera control in-game (e.g., we choose \texttt{U},\texttt{I},\texttt{O},\texttt{P} in Dark Souls, but any four keys would suffice), or to fixed-size mouse movements when only mouse look is available, as in Black Myth: Wukong.
For simplicity, we use the symbols \texttt{U},\texttt{I},\texttt{O},\texttt{P} throughout to refer abstractly to the four directional camera actions. The controller operates in a loop, combining (i) a finite-state machine (FSM) over high-level behavior, (ii) a pulse-based low-level camera controller, and (iii) a visual Memory Bank that biases decisions away from previously failed directions.

\subsubsection{Control Loop and Finite-State Machine}

At time-step $t$ the agent has an internal state
$s_t \in \mathcal{S}$, where $\mathcal{S}$ includes states such as
\textsc{Scan}, \textsc{Align}, \textsc{Advance}, a refinement state
(\textsc{Refine}), and recovery states
(\textsc{RecoverLocal}, \textsc{EscapeStuck}, \textsc{LoopBreaker}). Each step
proceeds as follows:
\begin{enumerate}
    \item \textbf{Screen capture and perception:}
    We capture the current frame $I_t$ via screen-grabbing and run the STP/MSTP
    model to obtain $\text{MSTP}_t$ and its sector $\sigma_t$.
    \item \textbf{Progress signals:}
    We maintain a ring buffer of recent frames (downsampled or grayscale) and
    compute progress features such as
    (i) change in MSTP box area,
    (ii) structural similarity index (SSIM) \cite{wang2004image} between $I_t$ and $I_{t-\Delta}$, and
    (iii) median optical flow magnitude. These statistics feed a
    progress meter that tracks how long the agent has gone without clear
    advancement.
    \item \textbf{FSM update:}
    Based on $(s_t, \text{MSTP}_t)$ and progress signals, we update
    $s_{t+1} = F(s_t, \text{MSTP}_t, \text{progress}_t)$.
    For example, if a stable MSTP is found after scanning, we transition
    \textsc{Scan} $\rightarrow$ \textsc{Align}; if forward motion persists
    but progress features stagnate, we escalate from \textsc{Advance} to a
    recovery state such as \textsc{RecoverLocal} or \textsc{EscapeStuck}; if
    repeated loop-like anchors are detected, we enter \textsc{LoopBreaker}.
    \item \textbf{Low-level control:}
    Given $s_{t+1}$ and $\text{MSTP}_t$, the controller outputs key pulses
    (camera + forward) as described next.
\end{enumerate}

\subsubsection{Pulse-Based Camera and Forward Control}

Let $\mathbf{c}_t = (u_t, v_t)$ be the MSTP center in pixel coordinates and $\mathbf{c}_0 = (u_0, v_0)$ the screen center. We define a normalized error vector
\[
    \mathbf{e}_t = \left(
        \frac{u_t - u_0}{W_{\text{screen}}},
        \frac{v_t - v_0}{H_{\text{screen}}}
    \right) = (e^x_t, e^y_t),
\]
To avoid rapid oscillations when the target is close to the screen center, we use a Schmitt-trigger style \cite{schmitt1938thermionic} dead zone with hysteresis: small errors inside an inner band do not trigger camera pulses, while a larger outer threshold is required to re-enter the \textsc{Align} state once the MSTP has been centered. Concretely, we declare
\begin{align*}
    \text{heading aligned} &\iff |e^x_t| < \epsilon_x,\\
    \text{pitch aligned} &\iff |e^y_t| < \epsilon_y.
\end{align*}

The camera controller generates brief pulses when the error exceeds threshold. In our implementation we realize this via a \emph{pulse controller} that maps error magnitude to a number of key pulses with a minimum inter-pulse gap. If $e^x_t$ or $e^y_t$ crosses its threshold we emit the corresponding directional key in $\{\texttt{U},\texttt{I},\texttt{O},\texttt{P}\}$ (otherwise $\varnothing$), and we set the number of taps to
$
    n_t = \mathrm{clip}\!\left(\big\lfloor k \|\mathbf{e}_t\|\big\rfloor, 0, n_{\max}\right),
$
each tap lasting $T_{\text{pulse}}$ and separated by at least $T_{\text{gap}}$ to avoid left-right jitter, thereby approximating a proportional controller.

Forward movement is gated by heading alignment via a Schmitt trigger. We
maintain a binary forward flag $f_t \in \{0,1\}$ and update
\[
    f_{t+1} =
    \begin{cases}
        1 & \text{if heading aligned for } \tau_{\text{on}} \text{ and in a
        forward-capable state}, \\
        0 & \text{if heading misaligned or in a recovery state}.
    \end{cases}
\]
When $f_{t+1}=1$ and $f_t=0$, we press and hold \texttt{W}; when
$f_{t+1}=0$ and $f_t=1$, we release \texttt{W}. The final action at
time-step $t$ is a combination of camera pulse (possibly $\varnothing$) and
forward toggle:
\[
    a_t = \bigl(a^{\text{cam}}_t, \Delta f_t\bigr),
    \quad \Delta f_t = f_{t+1} - f_t.
\]

\subsubsection{Visual Memory Bank}

To reduce repeated failures (e.g., repeatedly turning into a dead end or ``air-wall''), we maintain a visual memory bank that records past frames, directions chosen, and their outcomes. The bank has two partitions: \emph{quarantine} (recent entries waiting to be activated) and \emph{active} (entries that can bias decisions). This component is optional in our system and can be enabled or disabled at runtime. Each memory entry $m$ stores
\[
    m = \bigl(z, h, \sigma, t_{\text{dec}}, o\bigr),
\]
where:
\begin{itemize}
    \item $z = f_{\text{emb}}(I) \in \mathbb{R}^d$ is a frame embedding from the same feature extractor used for STP/MSTP retrieval \cite{xu2025adaptive},
    \item $h$ is a 64-bit perceptual hash of the frame, computed using a standard discrete cosine transform (DCT)-based image hashing scheme following prior work on perceptual image hashing \cite{monga2006perceptual, swaminathan2006robust},
    \item $\sigma \in \{1,\dots,K\}$ is the sector that was chosen at decision time,
    \item $t_{\text{dec}}$ is the decision timestamp,
    \item $o \in \{\text{GOOD},\text{BAD},\text{UNK}\}$ is an outcome label.
\end{itemize}
Intuitively, GOOD marks local views where the chosen sector led to a stable MSTP in front of the agent, BAD marks choices that quickly lost the MSTP (e.g., turning into a wall or dead end), and UNK is used for entries that have not yet accumulated enough evidence.

\paragraph{Insertion and promotion: }
Every $N$ frames, we consider inserting the current frame $I_t$ with its current ``preferred sector'' $\hat{\sigma}_t$ of the MSTP into quarantine. We only insert if $I_t$ is sufficiently novel:
\[
\min_{m' \in \mathcal{M}} \mathrm{Ham}(h_t, h_{m'}) > \delta_h
\ \text{and}\
\max_{m' \in \mathcal{M}} \cos(z_t, z_{m'}) < \delta_z .
\]

where $\mathcal{M}$ is the set of existing memories, $\text{Ham}$ is the Hamming distance between 64-bit pHashes, and $\cos$ is cosine similarity in embedding space. Perceptual hashes are designed so that visually similar images map to nearby binary codes that can be compared efficiently with Hamming distance while remaining robust to photometric changes \cite{monga2006perceptual, swaminathan2006robust}. Binary visual descriptors such as ORB are matched using Hamming distance for scalable visual retrieval and place recognition \cite{rublee2011orb}. The additional cosine threshold on $z_t$ prevents inserting frames that are well covered in the learned embedding space; cosine similarity is a standard choice for deep visual embeddings trained with metric-learning objectives \cite{schroff2015facenet}. New entries start with $o=\text{UNK}$ and a future activation time $t_{\text{act}} = t + T_{\text{quar}}$. When $t \ge t_{\text{act}}$, we move them from quarantine to active and add them to an approximate nearest-neighbor (ANN) index on $z$.

\paragraph{Decision association and outcome:}
When the controller commits to a sector decision $\sigma^\ast_t$ (e.g., when switching to \textsc{Advance} along that sector), we find the nearest quarantine entry $m$ whose frame time is within a small window before $t$ and set
$
    \sigma_m \gets \sigma^\ast_t, \quad t_{\text{dec},m} \gets t.
$
After a short delay $T_{\text{eval}}$ we assign an outcome label based on the stability of the MSTP selection. In our implementation, we compare the recorded decision box with the MSTP chosen at time $t_{\text{dec},m} + T_{\text{eval}}$ using intersection-over-union (IoU). If a MSTP is available and the IoU exceeds a threshold $\tau_{\text{IoU}}$, we set $o_m = \text{GOOD}$; if no MSTP is selected at that time or the IoU falls below $\tau_{\text{IoU}}$, we set $o_m = \text{BAD}$. The richer progress signals based on MSTP area, SSIM, and optical flow are used by the FSM to detect stagnation and trigger recovery states, but are not directly used to label memory outcomes.

\paragraph{Biasing candidate sectors:}
Before choosing an MSTP/sector at time $t$, we query the active bank with the current embedding $z_t$ and retrieve the $k$ nearest neighbors $\mathcal{N}_t$. For each candidate sector $\sigma$, we compute a penalty
\[
    \mathrm{pen}(\sigma \mid z_t)
    = \lambda \sum_{m \in \mathcal{N}_t}
      w(m,\sigma) \, \mathbb{I}[o_m = \text{BAD}],
\]
where $w(m,\sigma)$ encodes similarity, directional alignment, and recency. In our implementation $w(m,\sigma)$ is realized as the product of: (i) cosine similarity between $z_t$ and $z_m$,
(ii) a Gaussian kernel on sector distance (penalizing sectors close to
$\sigma_m$), and (iii) an exponential time decay (recent failures weigh more), with an additional multiplicative factor for BAD outcomes. When scoring MSTP candidates $\{b_i\}$ with sectors $\sigma(b_i)$, we adjust
$
    s_i^{\text{final}} = s_i - \mathrm{pen}(\sigma(b_i) \mid z_t),
$
and select the MSTP using $s_i^{\text{final}}$ instead of $s_i$.

Intuitively, when the current frame is visually similar to earlier states where turning into sector $\sigma$ led to a bad outcome, the memory bank reduces the effective score of candidates in sector $\sigma$, encouraging the controller to try alternative directions. This mechanism is purely frame-based and thus remains compatible with the screen-only constraint. In all our experiments the memory bank is optional and can be toggled on/off to study its contribution separately from the base perception and control stack. Full details are provided in the project repository.

\section{Experiments}
In this section we describe the experimental protocol used to evaluate our navigation agent in commercial 3D ARPG levels. We first detail the milestone-based setup and automation pipeline, then specify the routes, games, and agent variants in our pilot study.

\subsection{Experimental Setup}
To obtain reproducible, screen-only progress signals at the level of whole rooms and corridors, we evaluate our agent using a library of \emph{visual milestones}. Each milestone corresponds to a narrow band of camera poses around a designer-chosen target view (e.g., a door at the end of a hallway). During evaluation, the agent has no access to game coordinates, navigation meshes, or engine state; it must navigate purely from MSTP predictions, while we use template matching against these milestone views to detect when it has reached a desired location.

For each test level $\ell$ we first run an interactive capture tool while manually playing. The tool grabs the game window via screen capture and, whenever the user presses a capture hotkey, performs a short yaw sweep around the current view, optionally at several nearby pitch angles: it sends a sequence of small left/right camera pulses, waits briefly after each turn, and records the resulting frames as grayscale templates. All templates from these sweeps share a common group identifier and collectively represent a single logical milestone under slightly different headings, improving robustness to small viewpoint changes. Repeating this process at different locations and times yields a set of milestone groups for level $\ell$.

At evaluation time we load all templates for level $\ell$, group them by identifier, and sort the groups into a fixed milestone sequence $\mathcal{M}_\ell = (g_1, g_2, \dots, g_{K_\ell})$. The agent maintains an index into this sequence and always treats $g_j$ as the \emph{current target milestone}. Templates from already completed groups are ignored for the remainder of the run, and later milestones are never considered before all earlier groups have been completed or failed. A reset hotkey rewinds this index and clears visit history, allowing multiple runs per level.

Once automation is enabled, the agent periodically checks for milestone hits. Every $n$ frames it converts the current screen frame to grayscale, resizes it to each template in the current group $g_j$, and applies normalized cross-correlation template matching. If the maximum match score across templates in $g_j$ exceeds a fixed threshold, we declare that $g_j$ has been reached, mark the group as completed, and advance the target index to $g_{j+1}$. Between milestones, we log behavior as a sequence of segments. A new segment starts when automation is enabled or immediately after each completed milestone, and ends when the next milestone is reached, when the per-segment time limit is exceeded, or when automation is manually stopped. For each segment we record wall-clock start and end times, the start and end milestone identifiers, the termination reason (milestone reached, timeout, manual stop), the total number of frames and MSTP decisions, the number and cumulative duration of forward key presses, a histogram of camera pulses per direction, and dwell time in each FSM state. These logs support both coarse metrics (e.g., milestones reached per run, per-milestone success or failure) and fine-grained behavioral statistics (e.g., time spent scanning vs.\ advancing) aligned with the visual milestone sequence.

To evaluate end-to-end routes while preserving segment-level structure, we enforce a fixed time budget for each target milestone. If the agent does not reach $g_j$ within this budget, the current segment is terminated and $g_j$ is marked as a failure. During milestone capture we also record a game save at each milestone group (e.g., by copying the current save slot into a milestone-specific folder), and at evaluation time the agent uses these milestone saves to automatically reposition: after a timeout it drives back to the main menu, swaps the active save file for the one associated with the next milestone $g_{j+1}$, and then re-enters the game. This protocol lets us distinguish milestones reached autonomously from those that require a reset while still measuring screen-only navigation behavior across the entire, ordered path, and in practice supports fully automated, unattended testing runs.

\subsection{Pilot Experiment Settings}

\begin{table}[t]
  \centering
  \scriptsize
  \caption{Pilot study configuration. Game abbreviations: DS1 = Dark Souls I, DS3 = Dark Souls III, ER = Elden Ring, BMW = Black Myth: Wukong. Each route is decomposed into 6 milestones, yielding $K_\ell$ navigation segments per run.}
  \label{tab:pilot-settings}
  \begin{tabular}{llp{0.5\columnwidth}cc}
    \toprule
    Game & Route ID & Description & Runs & Milestones \\
    \midrule
    DS1 & DS1-UP & Undead Parish: from the dragon bridge up to the Gargoyle church & 10 & 6 \\
    DS1 & DS1-PW & Painted World of Ariamis: from the starting bridge to the staircase before the Undead Dragon & 10 & 6 \\
    DS3 & DS3-GA & Grand Archives: from the first bonfire to the second-floor landing & 10 & 6 \\
    DS3 & DS3-IR & Irithyll of the Boreal Valley: from the bonfire to the plaza before the second bonfire & 10 & 6 \\
    ER  & ER     & Raya Lucaria Academy: from the Schoolhouse Classroom up to the rooftop exit & 10 & 6 \\
    BMW & BMW    & Chapter~3 precept corridor: from the lower shrine toward the upper courtyard & 10 & 6 \\
    \bottomrule
  \end{tabular}
\end{table}

We conduct a pilot study across multiple commercial ARPGs to assess how far a screen-only, affordance-driven agent can navigate in real levels. We treat Dark Souls~I and Dark Souls~III as \emph{core} testbeds, since the STP/MSTP models were initially trained on these titles, and Elden Ring and Black Myth: Wukong (BMW) as \emph{generalization} titles used only for cross-game transfer tests. For each game we define one or more designer-specified routes from a bonfire (or equivalent checkpoint) to a boss gate or major landmark, and decompose each route into an ordered sequence of visual milestones, as summarized in Table~\ref{tab:pilot-settings}. To focus on purely visual guidance, all enemies along these routes are disabled, and we select paths that can be traversed using only basic walking controls without jumping or other special actions. In the core games (DS1, DS3) we evaluate two partial routes per title; in the generalization games (ER, BMW) we use a single route per title. Each level $\ell$ thus yields a milestone sequence $\mathcal{M}_\ell = (g_1, \dots, g_{K_\ell})$ with $K_\ell = 6$; we treat each $g_j$ as the target of one navigation segment, starting from the previous milestone (or the spawn point for $g_1$).

For every game-level pair and each agent configuration, we perform multiple full-route runs. At the start of a run, the avatar is placed at an initial starting point (before the first milestone) and the camera is aligned to match the first milestone group $g_1$; the agent then runs autonomously through the entire milestone sequence. If a target milestone $g_j$ is reached before its per-segment time budget expires, the agent immediately begins a new segment toward $g_{j+1}$ with a new time budget. If the time budget is exceeded, we terminate the current segment, mark $g_j$ as a failure, and use the corresponding milestone save for $g_{j+1}$ to automatically reposition the avatar at the next milestone before resuming automation. This protocol guarantees at least one complete pass over the prescribed route per run while distinguishing milestones reached autonomously from those requiring a reset, and it allows us to run large batches of experiments in a fully automated, screen-only fashion.

We evaluate three agent configurations. The \emph{Naive} variant directly steers toward the current MSTP with minimal recovery logic. The \emph{FSM-only} variant enables the full finite-state controller (SCAN, ALIGN, ADVANCE, REFINE, RECOVER\_LOCAL, ESCAPE\_STUCK, LOOP\_BREAKER) and progress-based recovery without the visual Memory Bank. The \emph{Full} variant uses the complete system with the Memory Bank. For core games (Dark Souls~I/III) we run two levels ten times; for generalization games (Elden Ring, Black Myth: Wukong) we use the same per-level run counts but only a single level per title. Table~\ref{tab:pilot-settings} summarizes the resulting experimental grid. Top-down schematics with the designer-specified paths and milestone annotations for the levels that are not mentioned in the main text are provided in Appendix~\ref{sec:route-maps}.

\section{Results}
We evaluate three agent variants (Naive, FSM-only, and Full including the Memory Bank) across four commercial ARPGs and multiple designer-specified routes, treating each level as a sequence of six visual milestones and six navigation segments. For every game-level-method combination we aggregate segment-level logs over ten full runs into milestone-level and level-level statistics, focusing on milestone success, timeout behavior, and temporal efficiency under our fixed per-segment time budget. Table~\ref{tab:overall-level-method} summarizes overall performance per game, level, and method in terms of a full-route success rate (RS) and milestone-level averages (MS), while Table~\ref{tab:per-milestone-success} reports per-milestone segment success rates for the Dark Souls~III and Black Myth: Wukong levels we analyze in more detail below.

\begin{table}[t]
  \centering
  \small
  \caption{Overall navigation performance per route and method. RS is the route success rate, the percentage of full runs that reach all six milestones; MS is the per-milestone segment success rate. Durations are reported in seconds.}
  \label{tab:overall-level-method}
  \begin{tabular}{llcccc}
    \toprule
    Route & Method & RS (\%) & MS (\%) & Seg. dur. (s) & Fwd time (s) \\
    \midrule
    DS3-GA   & Naive & 20 & $83.3 \pm 29.3$ & $84.9 \pm 56.8$  & $22.4 \pm 11.6$ \\
       & \textbf{FSM}   & 50 & \boldmath$88.3 \pm 16.8$ & $70.6 \pm 66.6$  & $17.1 \pm 13.1$ \\
       & Full  & 20 & $78.3 \pm 31.8$ & $99.0 \pm 100.0$ & $32.0 \pm 30.2$ \\
    DS3-IR   & Naive & 20 & $73.3 \pm 27.5$ & $117.9 \pm 59.9$ & $87.6 \pm 44.6$ \\
       & FSM   & 30 & $71.7 \pm 24.1$ & $137.4 \pm 85.7$ & $50.5 \pm 32.3$ \\
       & \textbf{Full}  & 20 & \boldmath$75.0 \pm 26.9$ & $103.3 \pm 65.0$ & $43.2 \pm 31.4$ \\
    \midrule
    DS1-UP & Naive & 0  & $40.0 \pm 33.2$ & $184.6 \pm 143.7$ & $20.2 \pm 41.0$ \\
     & FSM   & 0 & $55.0 \pm 27.5$ & $166.6 \pm 137.5$ & $22.2 \pm 32.9$ \\
     & \textbf{Full}  & 10 & \boldmath$58.3 \pm 26.7$ & $155.4 \pm 136.1$ & $27.7 \pm 41.1$ \\
    DS1-PW & Naive & 0  & $38.3 \pm 25.4$ & $200.3 \pm 130.3$ & $13.7 \pm 22.9$ \\
     & \textbf{FSM}   & 0  & \boldmath$48.3 \pm 38.0$ & $194.9 \pm 125.8$ & $84.0 \pm 64.5$ \\
     & Full  & 0  & $46.7 \pm 32.0$ & $198.9 \pm 127.2$ & $87.1 \pm 63.0$ \\
    \midrule
    BMW    & Naive & 0 & $61.7 \pm 41.0$ & $124.7 \pm 101.4$ & $73.2 \pm 64.8$ \\
        & \textbf{FSM}   & 40 & \boldmath$83.3 \pm 23.6$ & $136.7 \pm 46.9$  & $36.5 \pm 20.7$ \\
        & Full  & 20 & $75.0 \pm 35.5$ & $120.2 \pm 78.1$  & $47.7 \pm 29.9$ \\
    \midrule
    ER     & Naive & 0  & $27.0 \pm 25.6$ & $234.4 \pm 114.6$ & $184.1 \pm 118.3$ \\
         & FSM   & 0  & $38.3 \pm 31.3$ & $215.9 \pm 126.7$ & $74.4 \pm 61.1$ \\
         & \textbf{Full}  & 0  & \boldmath$52.0 \pm 24.2$ & $169.5 \pm 135.3$ & $68.8 \pm 59.5$ \\
    \bottomrule
  \end{tabular}
\end{table}

\subsection{Metrics and Reporting}
We treat each visual milestone as the target of one navigation segment and evaluate agents at three nested granularities: per-segment, per-milestone, and per-level. At the segment level we log the termination reason (milestone reached, timeout, or manual stop), wall-clock duration, number of frames and MSTP decisions, total forward keypress time, and dwell time in each FSM state. At the milestone level we compute a segment success rate (MS) as the fraction of segments targeting that milestone whose termination reason is reaching a milestone. At the level of a full route we also compute a route success rate (RS), defined as the fraction of complete runs in which the agent reaches all six milestones without timing out on any segment. Table~\ref{tab:per-milestone-success} reports these success rates as percentages. At the level of an entire route we aggregate milestone statistics into per-level metrics for each method as shown in Table~\ref{tab:overall-level-method}. For a given game-level-method combination we report the mean $\pm$ standard deviation of per-milestone success rates, segment durations, and forward times across the six milestones along the route. All runs use the same per-segment time budget, so longer durations reflect slow progress or repeated failures within a segment.

\begin{table}[t]
  \centering
  \small
  \caption{Per-milestone segment success rates for all routes. 
  $M_1$-$M_6$ denote successive milestone segments along each route. 
  Game abbreviations as in Table~\ref{tab:pilot-settings}.}
  \label{tab:per-milestone-success}
  \begin{tabular}{llcccccc}
    \toprule
    Route & Method & $M_1$ & $M_2$ & $M_3$ & $M_4$ & $M_5$ & $M_6$ \\
    \midrule
    DS3-GA  & Naive & 100\% & 100\% & 20\%  & 80\% & 100\% & 100\% \\
      & FSM   & 100\% & 100\% & 60\%  & 70\%  & 100\% & 100\% \\
      & Full  & 100\% & 100\% & 50\%  & 20\%   & 100\% & 100\% \\
    DS3-IR  & Naive & 100\% & 80\%  & 100\% & 60\%  & 80\%  & 20\%  \\
      & FSM   & 100\% & 70\%  & 60\%  & 70\%  & 100\% & 30\%  \\
      & Full  & 100\% & 100\% & 60\%  & 60\%  & 100\% & 30\%  \\
    \midrule
    DS1-UP  & Naive & 100\% & 10\% & 30\%  & 40\% & 60\% & 0\% \\
      & FSM   & 100\% & 30\% & 30\%  & 30\%  & 80\% & 60\% \\
      & Full  & 100\% & 70\% & 20\%  & 30\%   & 70\% & 60\% \\
    DS1-PW  & Naive & 60\% & 30\%  & 0\% & 30\%  & 30\%  & 80\%  \\
      & FSM   & 70\% & 40\%  & 0\%  & 90\%  & 0\% & 90\%  \\
      & Full  & 70\% & 50\% & 10\%  & 90\%  & 0\% & 60\%  \\
    \midrule
    BMW   & Naive & 100\% & 0\%   & 100\% & 20\%  & 50\%  & 100\% \\
       & FSM   & 100\% & 50\%  & 100\% & 100\% & 50\%  & 100\% \\
       & Full  & 100\% & 30\%  & 100\% & 100\% & 20\%  & 100\% \\
    \midrule
    ER    & Naive & 50\% & 0\%   & 20\% & 20\% & 70\% & 0\%  \\
        & FSM   & 70\% & 0\%   & 20\% & 50\% & 70\% & 20\% \\
        & Full  & 80\% & 30\%  & 40\% & 40\% & 90\% & 30\% \\
    \bottomrule
  \end{tabular}
\end{table}

\begin{figure}[t]
  \centering
  \includegraphics[width=\linewidth]{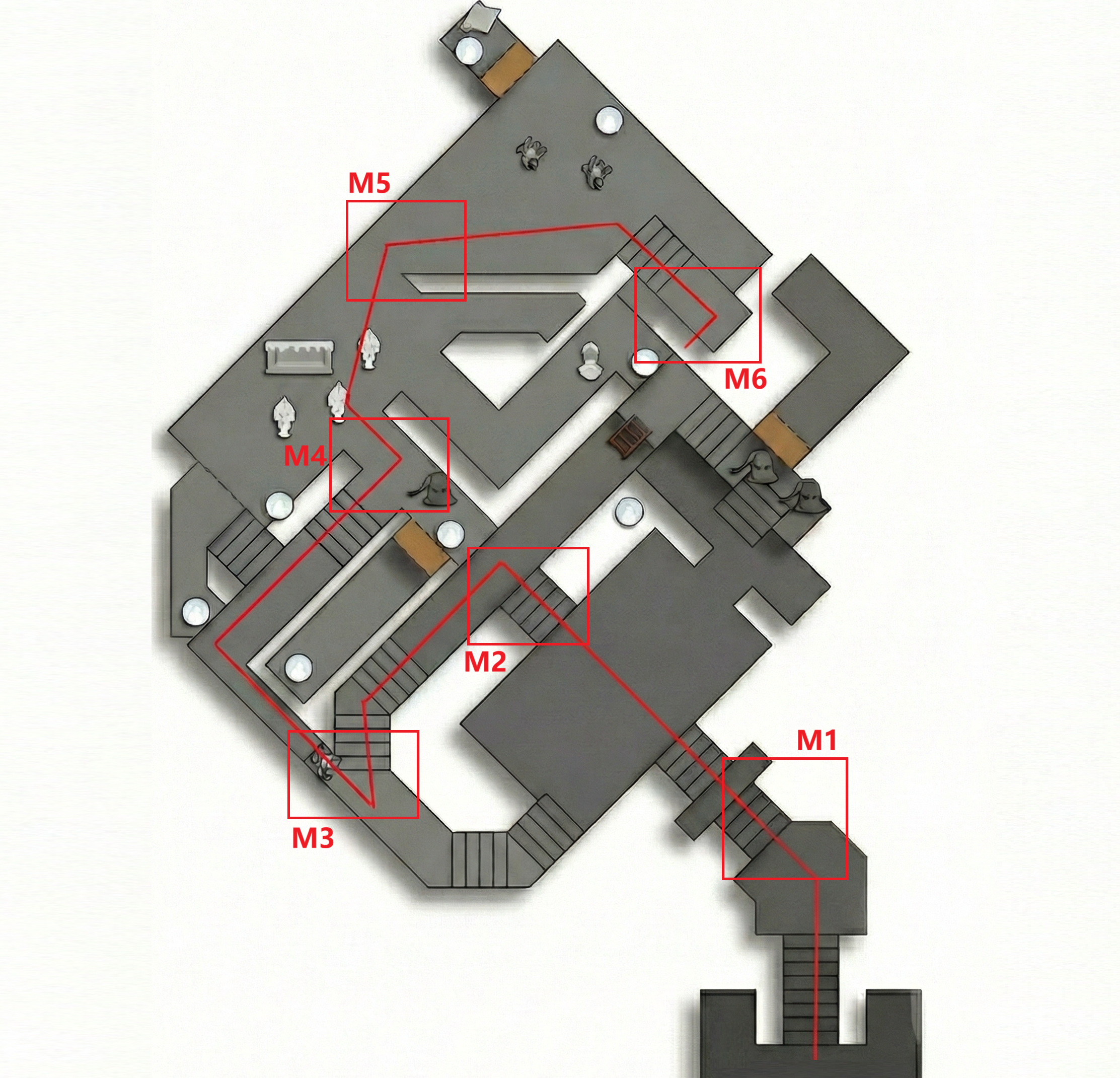}
  \caption{Top-down map of the Dark Souls~III Grand Archives test route. The red polyline shows the designer-specified path from the bonfire (bottom) toward the second-floor landing (top). Red boxes mark the six visual milestones $M_1$-$M_6$ used in our experiments; $M_2$ and $M_3$ correspond to the challenging mid-route transitions analyzed in
  Section~\ref{sec:case-study}.}
  \label{fig:ga-route}
\end{figure}

\subsection{Core Games: Dark Souls I and III}

In Dark Souls~III, where all three variants are fully evaluated on two routes, we see that screen-only navigation is feasible across a large fraction of the path but remains sensitive to method choice. Figure~\ref{fig:ga-route} summarizes the Grand Archives route used in our experiments, with milestones $M_1$-$M_6$ overlaid on a top-down schematic map. In Grand Archives (Table~\ref{tab:overall-level-method}), Naive and FSM both achieve high average milestone success (MS: $83.3 \pm 29.3\%$ and $88.3 \pm 16.8\%$), with FSM reducing mean segment duration from $84.9 \pm 56.8$\,s to $70.6 \pm 66.6$\,s and cutting average forward time from $22.4 \pm 11.6$\,s to $17.1 \pm 13.1$\,s. At the level of full routes, however, RS increases from 20\% for Naive to 50\% for FSM, while the Full agent reaches only 20\% despite comparable MS ($78.3 \pm 31.8\%$) and longer durations ($99.0 \pm 100.0$\,s, $32.0 \pm 30.2$\,s forward time), indicating that naive visual memory can sometimes hurt robustness in this visually dense level. Per-milestone success in Table~\ref{tab:per-milestone-success} shows that all methods stably solve the first two segments ($M_1$ and $M_2$), but Naive still struggles on the third segment ($M_3$, $20\%$ success) whereas FSM and Full restore this to $60\%$ and $50\%$, respectively. On the subsequent segment ($M_4$), Naive and FSM achieve similarly high success (around $70$-$80\%$), suggesting that the main advantage of the FSM on this route is localizing the difficult transition at $M_3$ rather than consistently outperforming Naive on all milestones. On the Irithyll route the pattern is more mixed. Naive attains an average milestone success of $73.3 \pm 27.5\%$ with relatively long durations ($117.9 \pm 59.9$\,s) and high forward time ($87.6 \pm 44.6$\,s), reflecting sustained forward motion and occasional stalls. FSM maintains a comparable success rate ($71.7 \pm 24.1\%$) but redistributes effort, running slightly longer on average ($137.4 \pm 85.7$\,s) while using much less forward time ($50.5 \pm 32.3$\,s), suggesting more scanning and recovery. The Full agent slightly improves success to $75.0 \pm 26.9\%$ and shortens duration to $103.3 \pm 65.0$\,s and forward time to $43.2 \pm 31.4$\,s, indicating that, unlike in Grand Archives, memory provides a modest net benefit on this route. Table~\ref{tab:per-milestone-success} confirms that all three methods reliably clear the earliest segment ($M_1$), while success diverges on mid-route segments ($M_2$-$M_4$), where the Full agent often matches or exceeds both Naive and FSM.

On Dark Souls~I, both Undead Parish (DS1-UP) and Painted World of Ariamis (DS1-PW) are noticeably harder than the Dark Souls~III routes, with detailed annotated routes shown in
Appendix~\ref{sec:route-maps}. In Undead Parish (Table~\ref{tab:overall-level-method}), Naive attains only about 40\% average milestone success and never completes a full six-milestone run, while the FSM and Full variants raise MS into the mid-50\% range and slightly shorten mean segment duration. Only the Full agent ever finishes the entire route (RS = 10\%), and Table~\ref{tab:per-milestone-success} shows why: all methods reliably clear the initial straight segment ($M_1$), but they struggle with the cramped approach and castle interior. Naive almost never reaches the final church milestone ($M_6$), whereas FSM and Full recover some stability there (60\%) and the Memory Bank in particular helps on the second milestone ($M_2$ climbs from 30\% to 70\%), though at the cost of mixed changes on earlier turns. Painted World is harsher still: none of the three agents ever complete the full route (RS = 0), and MS remains below 50\% for all methods despite modest gains from Naive to FSM/Full. The per-milestone breakdown highlights persistent failures around $M_3$ and $M_5$, which require sharp camera reorientation, narrow walkways without railings, and multi-floor transitions-situations where our 2D STP/MSTP model has little notion of vertical structure or fall risk. These trends align with earlier observations that Dark Souls~I is inherently the hardest title for our visual affordance model, with older, noisier visuals and fewer training examples than later entries, and they underline that our controller's benefits are bounded by the quality and coverage of the underlying affordance predictions.

\subsection{Transfer Games: Black Myth and Elden Ring}

Black Myth: Wukong provides the first transfer target with different lighting, character scale, and encounter pacing. On its Chapter~3 precept-corridor route, Table~\ref{tab:overall-level-method} shows Naive achieves an average milestone success MS of $61.7 \pm 41.0\%$, with long and highly variable segment durations ($124.7 \pm 101.4$\,s) and the highest forward time among the three methods ($73.2 \pm 64.8$\,s), and it never completes the full route (RS = 0\%). FSM substantially improves milestone success to $83.3 \pm 23.6\%$ and raises RS to 40\%, though with slightly longer mean duration ($136.7 \pm 46.9$\,s), while roughly halving the forward time to $36.5 \pm 20.7$\,s. The Full agent lies in between, with MS $75.0 \pm 35.5\%$, RS 20\%, $120.2 \pm 78.1$\,s mean duration, and $47.7 \pm 29.9$\,s forward time, suggesting that visual memory introduces additional variability without consistently improving robustness on this unseen title. Per-milestone results in Table~\ref{tab:per-milestone-success} reveal that all three methods reliably complete the first ($M_1$) and final ($M_6$) segments in the route, but they diverge sharply on the second and fifth segments. Naive fails $M_2$ entirely ($0\%$ success) and only completes $M_5$ half of the time, while FSM raises success in $M_2$ and $M_5$ to $50\%$, and the Full agent reaches $30\%$ and $20\%$. These results highlight both the stability of the underlying controller across games and the sensitivity of performance to level-specific visual challenges.

Elden Ring gives the hardest transfer setting. We use a Raya Lucaria Academy route from the Schoolhouse Classroom site of grace up through the interior loop and balconies to the rooftop exit near the Red Wolf arena (Figure~\ref{fig:er-route}), with milestones that cross narrow walkways, sharp turns, and several floor changes where a small mistake drops the avatar back to earlier areas. As summarized in Tables~\ref{tab:overall-level-method} and~\ref{tab:per-milestone-success}, all three variants achieve low milestone success (MS $27$–$52\%$) and never complete the full route (RS = 0\%). The Naive agent spends most of its time pushing into walls, breakable props, and railings (forward time $184.1 \pm 118.3$\,s), and sometimes gets trapped in corners that a human would escape with a short jump. The FSM controller cuts forward time by more than half and slightly raises MS to $38.3 \pm 31.3\%$, but still fails at tight corners and multi-storey turns. The Full agent does best here (MS $52.0 \pm 24.2\%$, shorter durations and forward time), showing that visual memory can help avoid re-entering the same bad corridors, yet the jump-only shortcuts, dense destructible clutter, and vertical layout still keep our non-jumping controller from reliably finishing the route.

\begin{figure}[t]
  \centering
  \includegraphics[width=\linewidth]{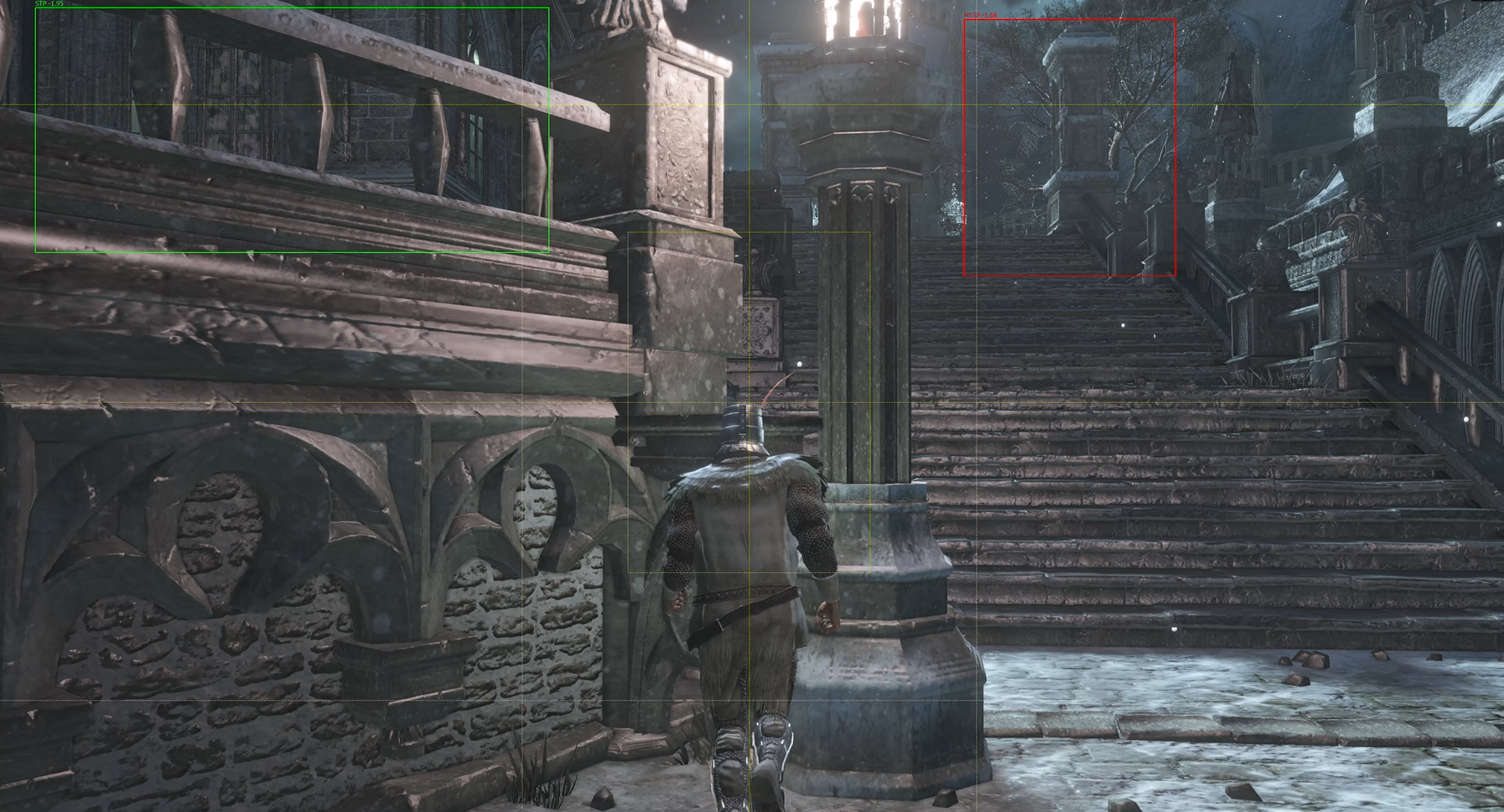}
  \caption{Irithyll $M_3$: the MSTP correctly highlights the upper landing, but the avatar gets stuck on local geometry.}
  \label{fig:ir-m3-stuck}
\end{figure}

\begin{figure}[t]
  \centering
  \includegraphics[width=\linewidth]{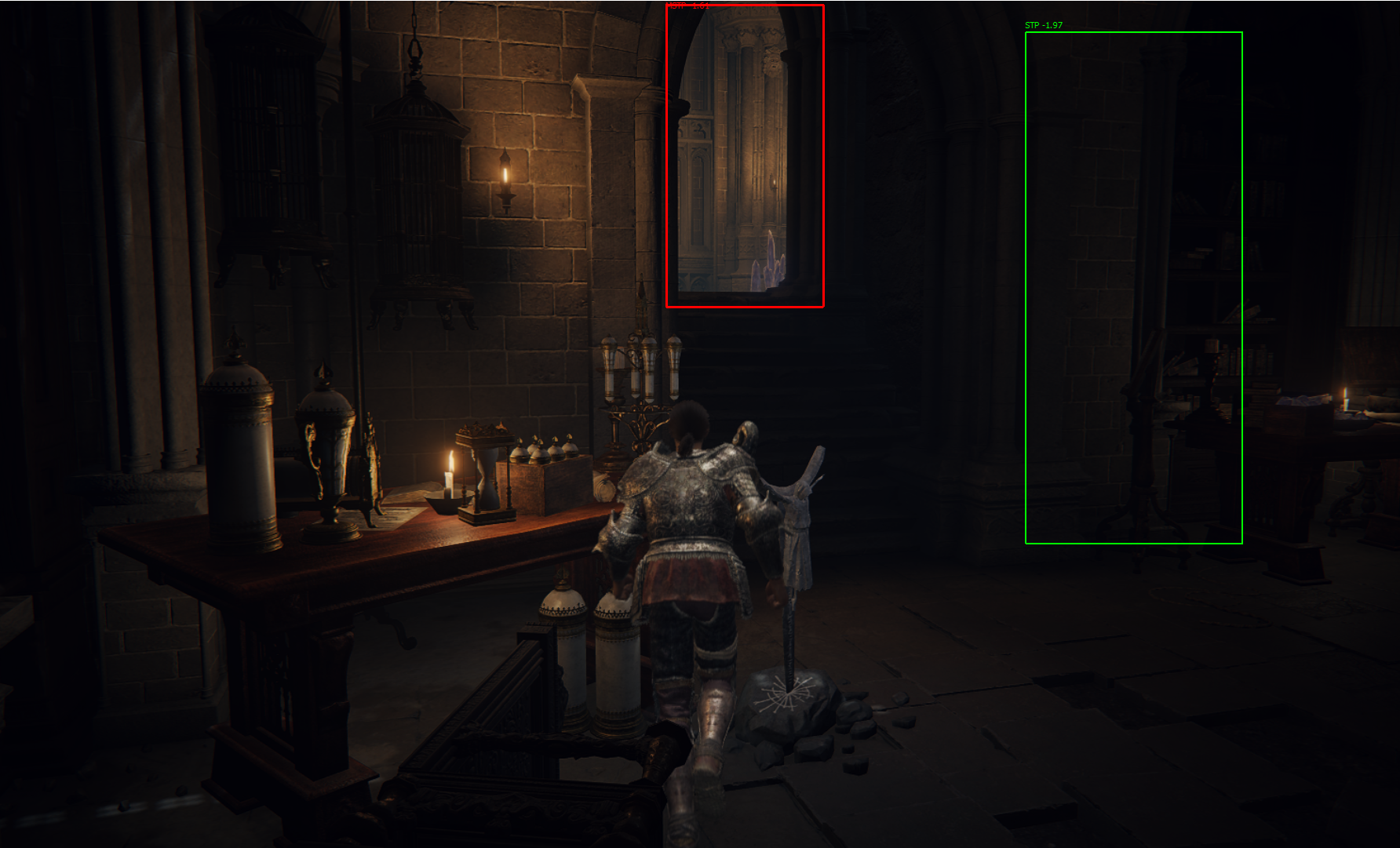}
  \caption{Raya Lucaria Academy from $M_1$ to $M_2$: the doorway is correctly detected as an MSTP, but the avatar gets stuck between a breakable table and an effigy.}
  \label{fig:er-rla-doorway}
\end{figure}

\begin{figure}[t]
  \centering
  \includegraphics[width=\linewidth]{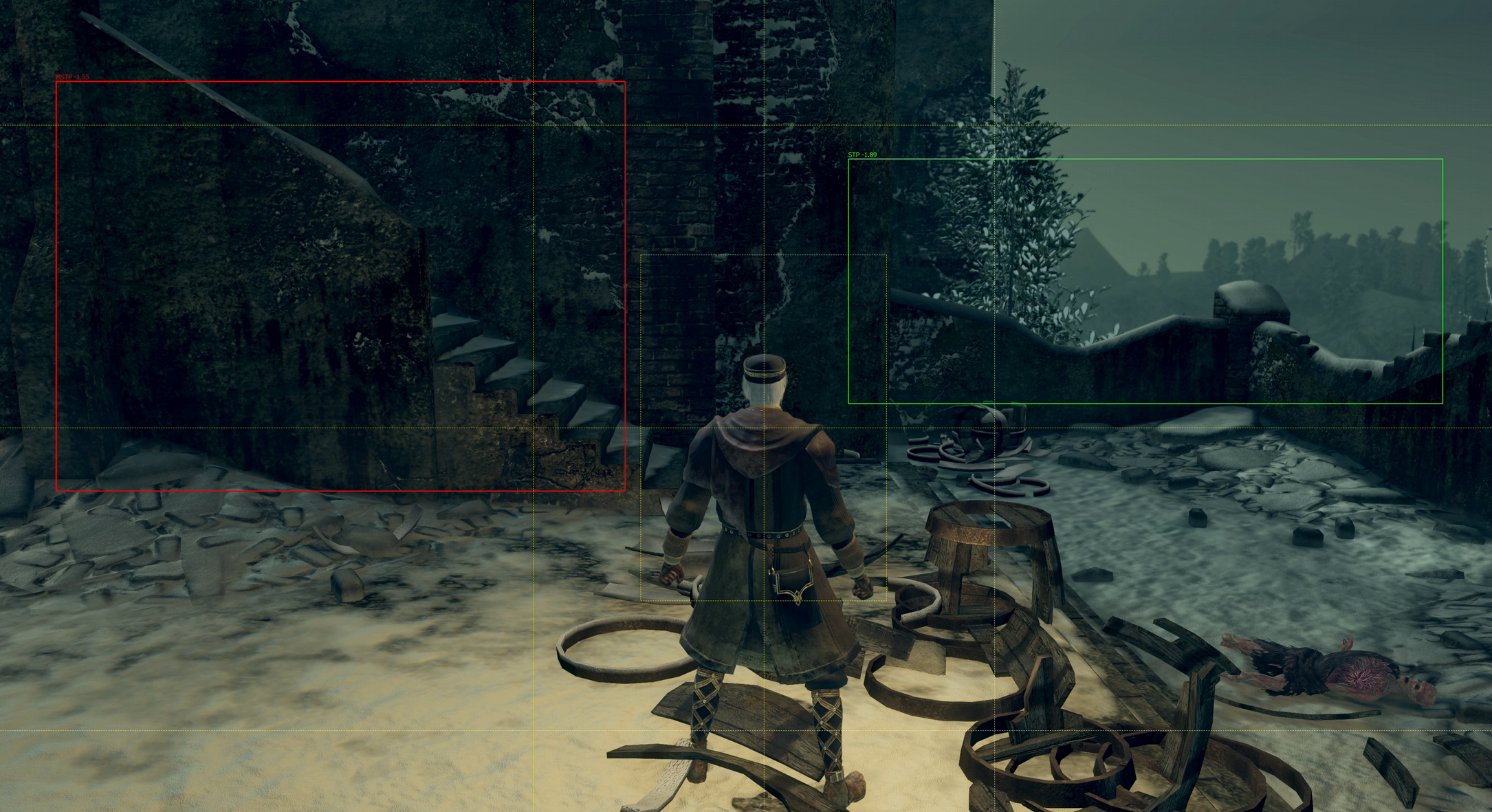}
  \caption{Painted World from $M_2$ to $M_3$: the stairs are correctly detected as an MSTP, but require a narrow, oblique approach angle that is easy to miss from the avatar's starting pose.}
  \label{fig:ds1-pw-stairs}
\end{figure}

\subsection{Case Study}
\label{sec:case-study}

Beyond aggregate metrics, a few concrete situations highlight both what our agent can do and where a purely screen-only affordance model breaks down. Figure~\ref{fig:ir-m3-stuck} shows Irithyll milestone $M_3$, a staircase bend where the correct transition is visually clear: the MSTP detector consistently fires on the upper landing, and the FSM steers the avatar straight toward it. However, the character repeatedly collides with the corner geometry and gets caught on the lamp post and step edges; progress metrics show long \textsc{Advance} dwell time with high forward key usage but almost no optical flow. Similarly, Figure~\ref{fig:er-rla-doorway} shows that, despite the correct MSTP being identified, the agent's approach has it ``stuck'' between an effigy and a breakable table, whereas a human player would choose to walk around or even roll through the table. A closely related failure appears in the Painted World route between milestones $M_2$ and $M_3$ (Figure~\ref{fig:ds1-pw-stairs}). Here the stairs that continue the critical path are correctly marked as the MSTP on the far left, but they are narrow, partially occluded, and must be mounted from a very specific angle. From the agent's typical facing direction, even small heading errors cause the avatar to scrape along the wall or slide off the step edge, so the controller often fails to commit to the climb at all. This explains why this route almost always breaks at $M_3$ in Table~\ref{tab:per-milestone-success}, despite apparently correct affordance predictions. These examples together illustrate that even when MSTPs are semantically correct, the controller still lacks the fine-grained world understanding and local collision reasoning needed to negotiate tight geometry and staircase-like structures; similar issues show up in Elden Ring, where breakable props frequently catch the avatar and turn small contact errors into full stalls. ``Seeing'' the goal is not the same as being able to traverse the space that leads there, and future agents will need shallow planning and physical reasoning in addition to per-frame affordance cues.

\begin{figure}[t]
  \centering
  \includegraphics[width=\linewidth]{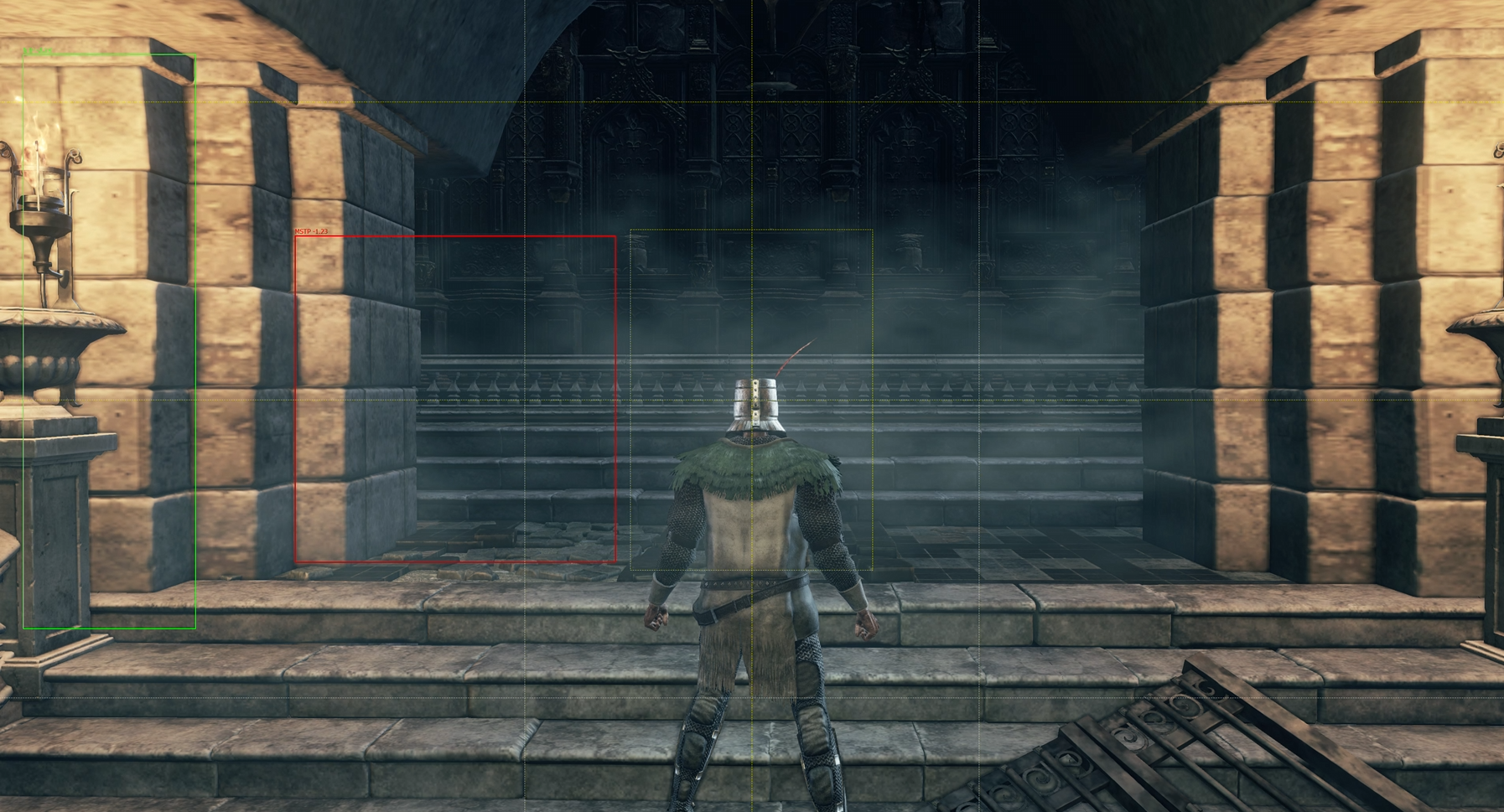}
  \caption{Grand Archives milestone $M_2$: a near-symmetric scene where both the data and the route favor turning left.}
  \label{fig:ga-m2-bias}
\end{figure}

\begin{figure}[t]
  \centering
  \includegraphics[width=\linewidth]{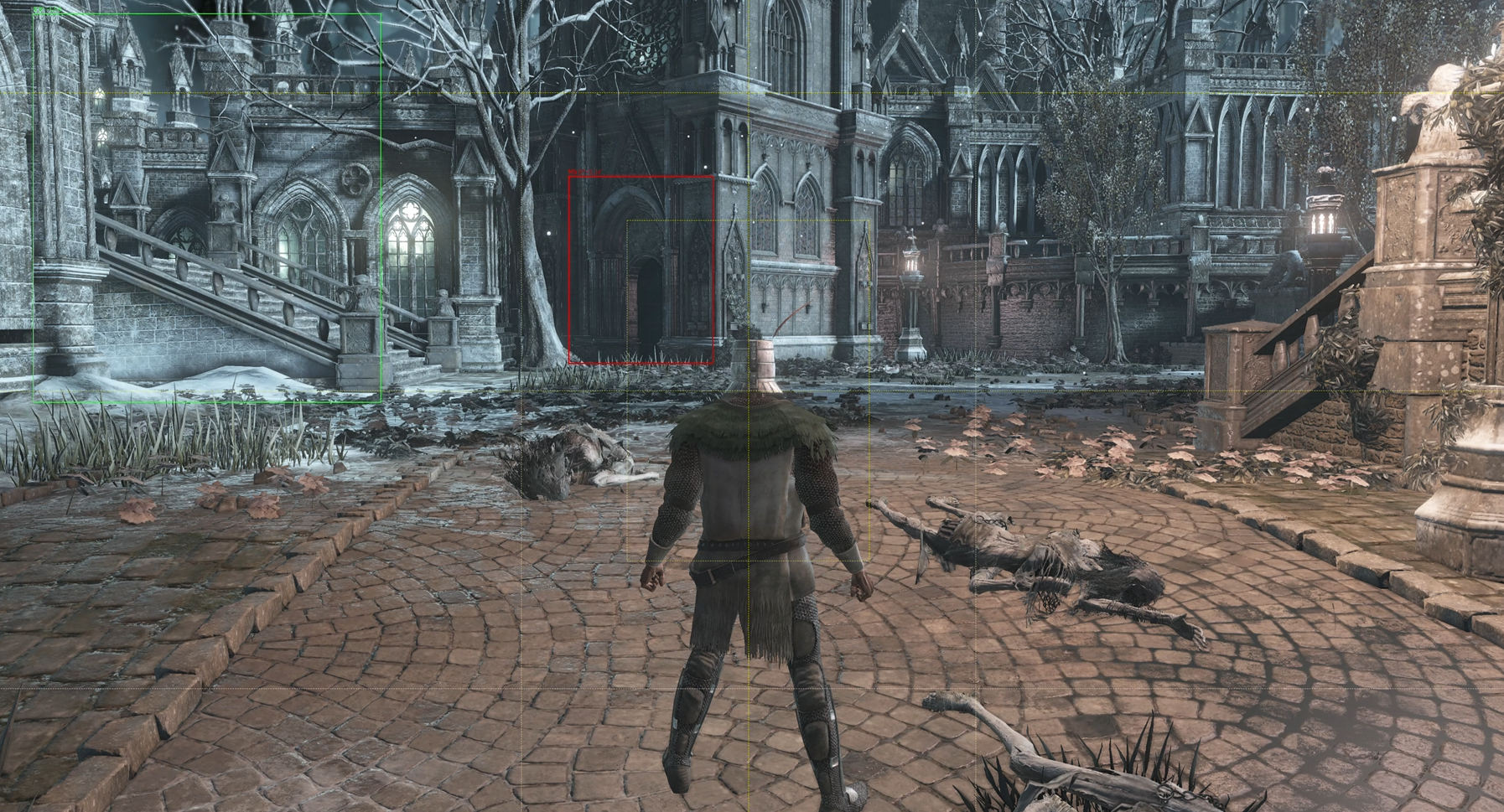}
  \caption{Irithyll milestone $M_6$: a visually similar symmetry, but here the intended route is the darker right-hand doorway that the model often overlooks.}
  \label{fig:ir-m6-bias}
\end{figure}

Our results also expose dataset-driven biases in the visual affordance model. As indicated in Figure~\ref{fig:ga-route}, Grand Archives milestone $M_2$ (Figure~\ref{fig:ga-m2-bias}) and Irithyll milestone $M_6$ (Figure~\ref{fig:ir-m6-bias}) both feature layouts where the walkable space is roughly mirror-distributed to the left and right of the avatar. In Grand Archives the designer path actually goes left, so the model's strong preference for left-hand STPs is harmless or even helpful. In Irithyll, however, the intended route turns into the darker right-hand route; the detector and selector nevertheless often lock onto the brighter, more open left facade. This left-leaning bias, inherited from our training corpus of single-frame STP annotations, explains why DS3-GA's mid-route segments remain solvable while DS3-IR's final segment ($M_6$) sees frequent failures, and it underlines that any attempt at ``general'' visual navigation must reason explicitly about training-distribution asymmetries rather than assuming salience detectors are neutral.

The behavior of the Full agent around Irithyll $M_6$ further clarifies the limits of our simple Memory Bank. In principle, repeatedly failing the left-hand turn should cause visual memories of that view to be marked as ``bad'' and down-weight future decisions in the same sector. In practice, Figure~\ref{fig:ir-m6-bias} shows that the correct doorway seldom appears as a strong STP candidate at all; the detector focuses on the bright staircase and decorative front door, while the right entrance is partially occluded by the avatar and environmental clutter. Because the Memory Bank can only reweight existing STP candidates rather than propose new ones, it mostly oscillates between similarly biased options and cannot systematically redirect the agent toward the unseen right-hand exit. This is consistent with the observations from the original paper, and matches our quantitative results, where memory slightly reshapes behavior but does not rescue difficult milestones, and suggests that effective visual memory must operate at the level of learned spatial hypotheses or topological maps instead of acting as a shallow penalty term on individual detections.

\begin{figure}[t]
  \centering
  \includegraphics[width=\linewidth]{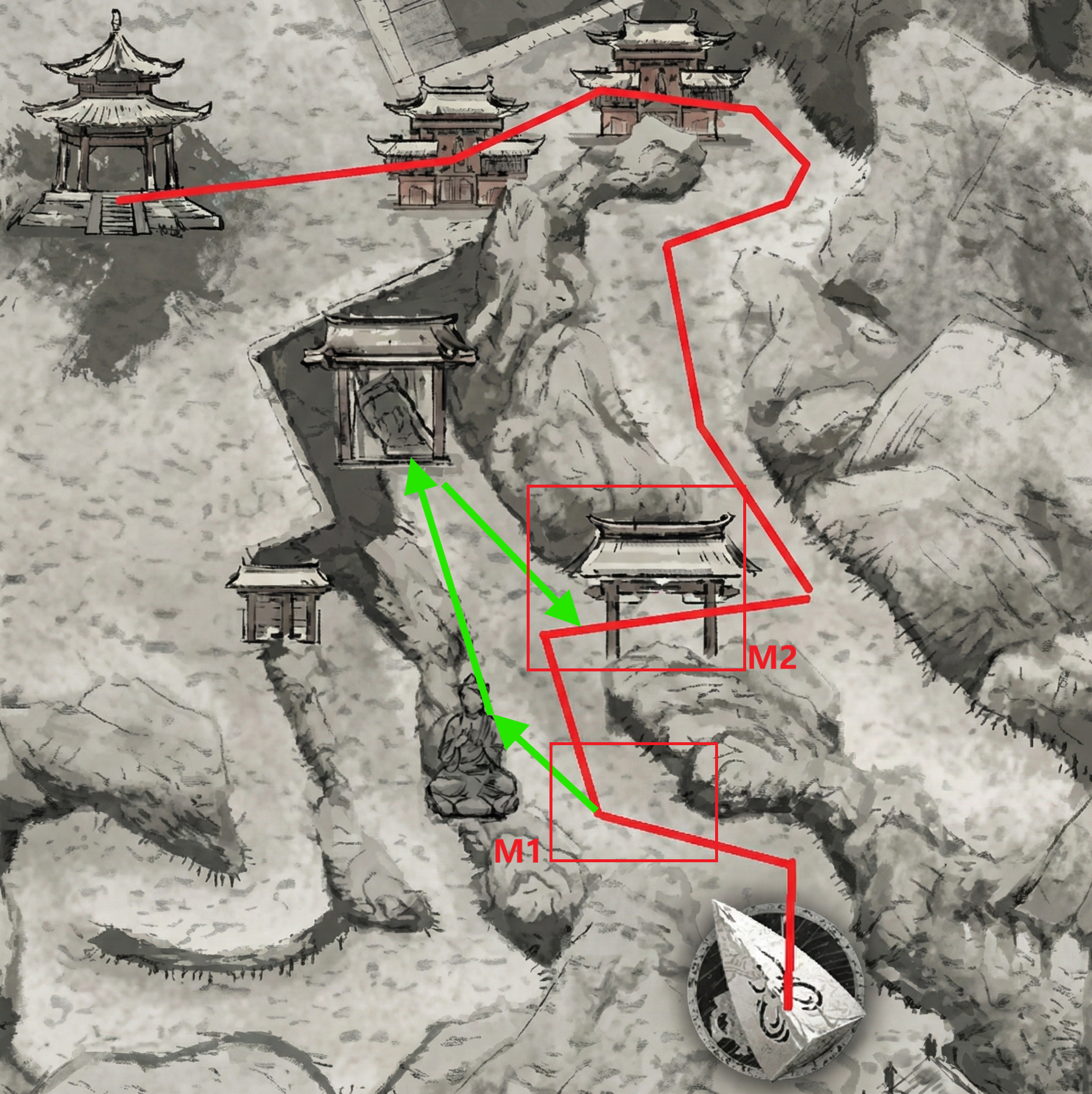}
  \caption{Black Myth: Wukong route. The red polyline shows the evaluation path and milestones $M_1$ and $M_2$; green arrows sketch the routes that players and our agents are visually drawn to first, toward left and upper side shrines, before noticing the narrow main-path gap near $M_2$.}
  \label{fig:bmw-route}
\end{figure}

A complementary perspective comes from our Black Myth: Wukong route, shown in Figure~\ref{fig:bmw-route}. Here the red polyline indicates the evaluation route from the starting shrine toward the upper courtyard, with milestones $M_1$ and $M_2$ marked in red. The green arrows, however, sketch the trajectories that human players often take: when standing at $M_1$, players are primarily drawn toward the visually prominent side shrines to the left and above, while the narrow main-path gap around $M_2$ is partially occluded by foreground rock and architecture. Players typically explore the upper dead end before noticing the subtle opening that continues the critical path. Our agents mirror this behavior. Naive almost always commits to the salient left and upper structures and rarely turns far enough to bring the $M_2$ gap into view, which explains its near-zero success on $M_2$ in Table~\ref{tab:per-milestone-success}. The FSM variant fares better: when it becomes stuck against the cliff or shrine geometry, RECOVER\_LOCAL and SCAN phases trigger additional sweeps that occasionally reveal the hidden opening, allowing it to progress to $M_2$. This case illustrates that the agent's mistakes are not irrational, nor do they imply bad level design-on the contrary, they are aligned with the designer's intention to use visual cues to encourage optional exploration. What it really exposes is a limitation of our evaluation objective: treating navigation as ``reach the main goal as fast as possible'' ignores that real players juggle curiosity, loot, and narrative payoffs alongside forward progress. To build agents that truly play these games rather than merely path-find through them, a visual controller like ours will need to be coupled with higher-level reasoning about objectives and value. At the same time, this example highlights a promising use case for our current system: as an automated probe for level-design intent, revealing where secondary attractions might drive visual guidance away from the intended main route.

\begin{figure}[t]
  \centering
  \includegraphics[width=\linewidth]{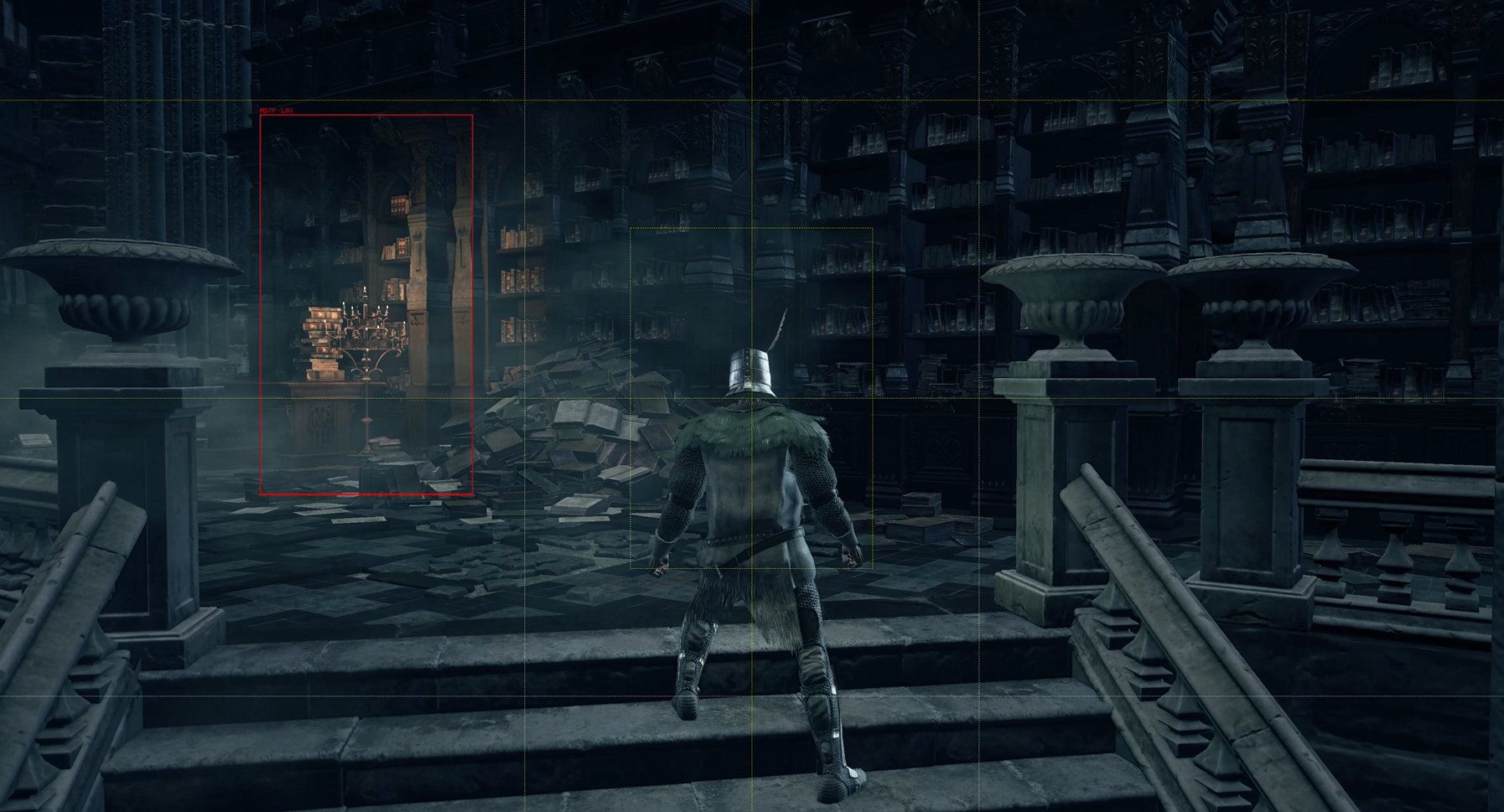}
  \caption{Grand Archives milestone $M_3$: decorative shelves and lighting dominate the frame, drawing the MSTP away from the right-hand turn that actually continues the route.}
  \label{fig:ga-m3-hidden}
\end{figure}

Finally, Grand Archives milestone $M_3$ (Figure~\ref{fig:ga-m3-hidden}) illustrates a complementary failure mode driven by misleading salience and hidden structure. The intended transition is a sharp right turn behind the camera into a side corridor, but the visible frame is dominated by richly detailed shelves, glowing candles, and a bright opening between bookcases. Our model confidently selects this visually striking patch as the MSTP, and the agent walks straight into a dead end or loops in front of the shelves, repeatedly timing out despite many successful detections. Here the true transition is only weakly implied in the current view, and the most visually distinctive region is not a traversable STP at all. The FSM sometimes mitigates this behavior: by inserting extra SCAN and RECOVER\_LOCAL phases, it can occasionally break out of the loop and reorient, but it cannot systematically fix it, and the Memory Bank offers little additional help because it can only reweight the same misleading candidates. Together with other examples, this makes clear that our failures are structural rather than accidental: as long as perception is limited to per-frame salience and local geometry, the agent will systematically confuse ``visually interesting'' with ``actually connected'' and lack a representation of turns that temporarily leave the camera frustum. Any future visual-only navigation system will therefore need to couple affordance prediction with some notion of long-range connectivity and world dynamics, whether through video-based world models, SLAM mapping, or learned topological graphs, rather than relying on pointwise STP detections alone.

\section{Conclusion}

In this paper we revisited visual affordance-driven navigation in modern 3D ARPGs from a systems and evaluation perspective. Building on an existing frame-based affordance model, we implemented a simple but game-agnostic controller that maps MSTP predictions to discrete camera and movement actions, and we introduced a reusable screen-only evaluation protocol based on designer-defined visual milestones. Using this protocol, we ran a pilot study across four commercial titles and six partial routes, comparing three agent variants and analyzing their behavior both through aggregate metrics and detailed case studies. Our results show that the FSM can offer meaningful robustness gains on specific segments, but that it does not consistently dominate the Naive baseline, and that the current memory mechanism often provides little benefit or even degrades performance. At the same time, the experiments show that nontrivial progress toward designer-specified goals is possible with purely visual input, and they provide a concrete benchmark and set of baselines for future work on screen-only navigation agents.

Our study has several important limitations. First, our controller is tightly coupled to a particular visual affordance model and its original training data, so our results are constrained by that model's coverage, biases, and failure modes, and we do not yet provide model-independent baselines that use different perceptual front-ends. This coupling also limits the conceptual novelty of the controller: some of the hardest parts of the problem, like detecting candidate transition points and ranking them, are inherited rather than learned in this work. Second, the empirical scope of our pilot is deliberately modest: six hand-picked routes across four games, with a single milestone configuration per route and a fixed time budget, are sufficient for case-study analysis but not for broad statistical claims about game or level diversity. These constraints mean that our contributions should be read as defining a task and exposing representative behaviors, rather than solving the general problem.

These limitations suggest several directions for future work. With an explicit, reproducible evaluation framework in place, we can now decouple the benchmark from any particular affordance backbone and explore alternative perception-control stacks: end-to-end visual RL, hierarchical policies that operate over learned topological maps, or hybrid approaches that combine short-horizon control with model-based rollouts and planning. In parallel, we see our affordance model not only as a navigation module but also as a promising tool for level design and analysis: our case studies already indicate that it is sensitive to designer-intended visual guidance and to dataset-driven biases. A natural next step is therefore to embed this controller into interactive HCI tools for designers and testers, supported by user studies that compare agent traces, designer intent, and player behavior. In both the ``agent'' and ``assistant'' directions, a key goal will be to scale up: more games, more routes, and larger training and evaluation corpora that move from pilot experiments toward practically useful testing workflows.

Stepping back, our findings underscore how far we still are from general-purpose navigation agents for contemporary, realistic 3D RPGs that rely on visual input alone. Such agents must not only identify ``where to go'' in each frame, but also reason about 3D geometry, physics, occlusion, and intermediate subgoals over extended horizons, all under severe partial observability. Our results also suggest that visual affordances by themselves are unlikely to support fully autonomous play: without some representation of world structure and long-range connectivity, a system will repeatedly confuse what is visually salient with what is actually traversable or important. Nevertheless, we believe that visual affordance models occupy a valuable niche. As assistive tools for designers and QA, they can act as automated probes of visual guidance, highlighting where level geometry, lighting, and decoration systematically steer agents (and likely players) away from intended routes. We hope that our benchmark, controller, and analysis will help catalyze progress along both fronts: toward stronger game navigation agents that integrate perception with world modeling and reasoning, and toward practical visual-affordance assistants that support the design and testing of complex 3D game worlds.

\bibliographystyle{ACM-Reference-Format}
\bibliography{references}

\clearpage
\appendix
\begin{figure}[!t]
  \centering
  \includegraphics[width=\linewidth]{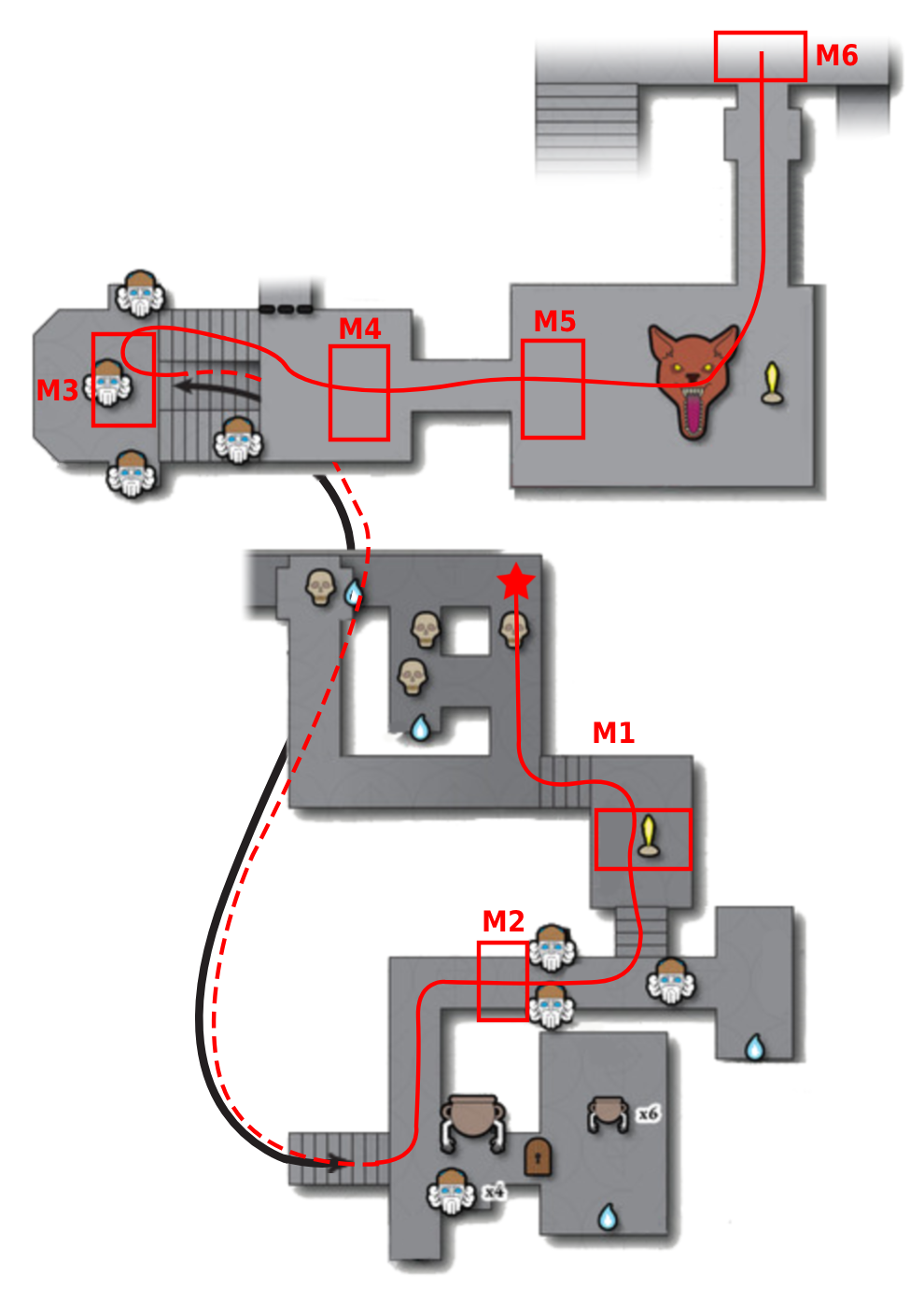}
  \caption{Elden Ring Raya Lucaria Academy route (ER). The red polyline shows the evaluation path from the Classroom site of grace up through the interior loop and balconies to the rooftop exit; red boxes mark milestones $M_1$-$M_6$.}
  \label{fig:er-route}
\end{figure}

\begin{figure}[!t]
  \centering
  \includegraphics[width=\linewidth]{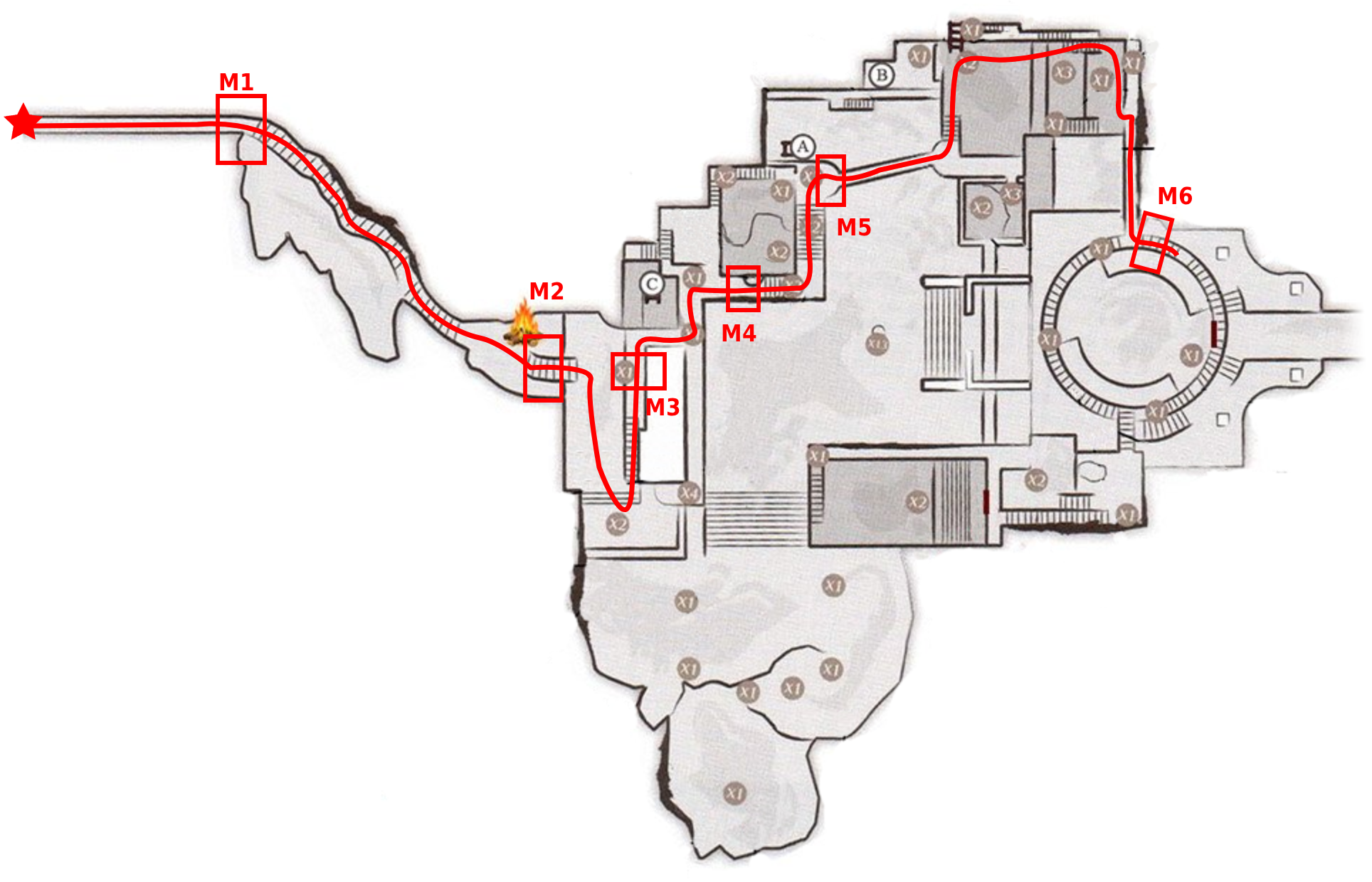}
  \caption{Dark Souls~I Painted World of Ariamis (DS1-PW) route. The polyline and rectangles indicate the evaluation path and milestones $M_1$-$M_6$.}
  \label{fig:ds1-pw-route}
\end{figure}

\begin{figure}[!t]
  \centering
  \includegraphics[width=\linewidth]{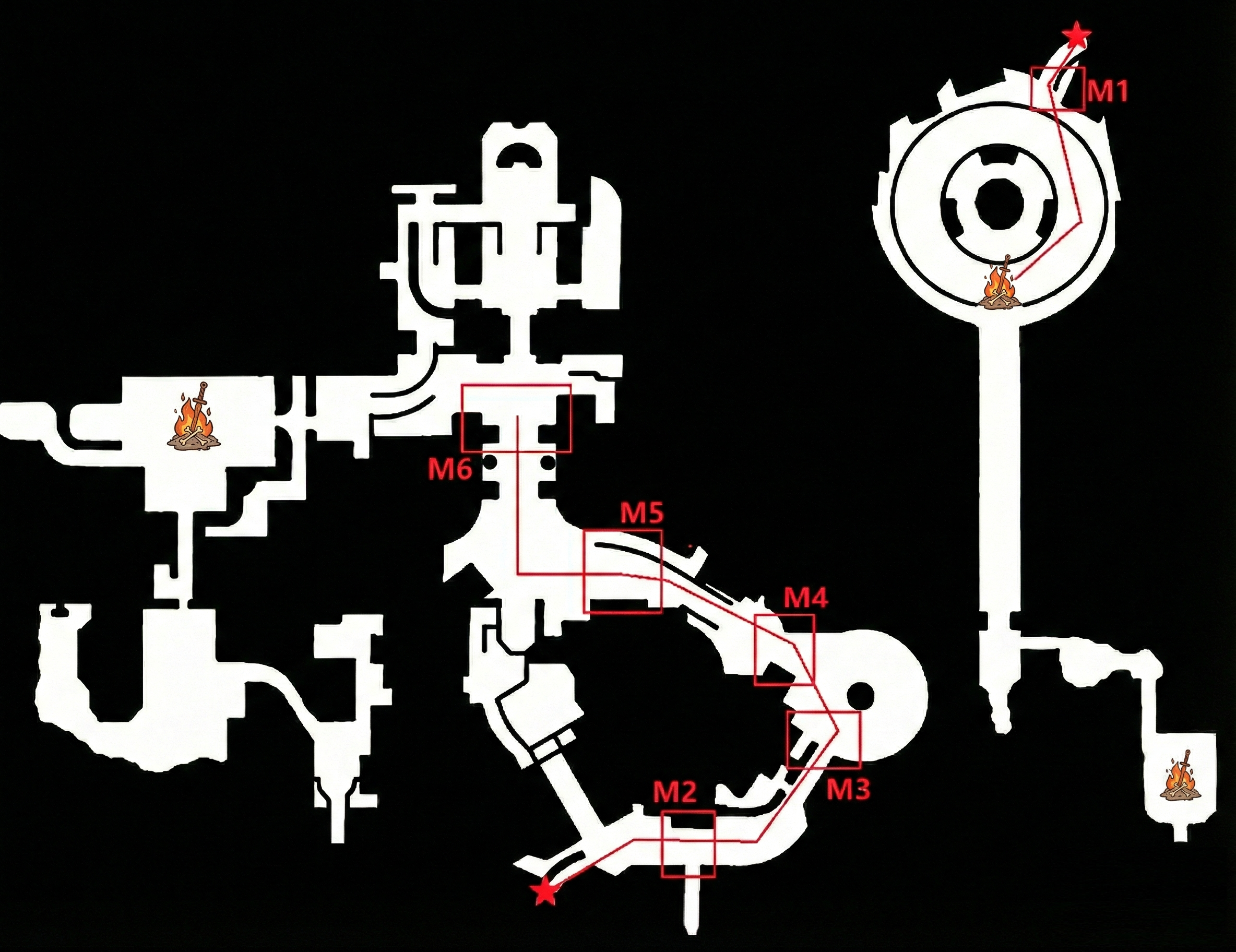}
  \caption{Dark Souls~III Irithyll of the Boreal Valley (DS3-IR) route. The red polyline shows the evaluation path from the bonfire toward the cathedral approach; red boxes mark the six visual milestones $M_1$-$M_6$ used in our experiments.}
  \label{fig:ds3-ir-route}
\end{figure}

\begin{figure}[!t]
  \centering
  \includegraphics[width=\linewidth]{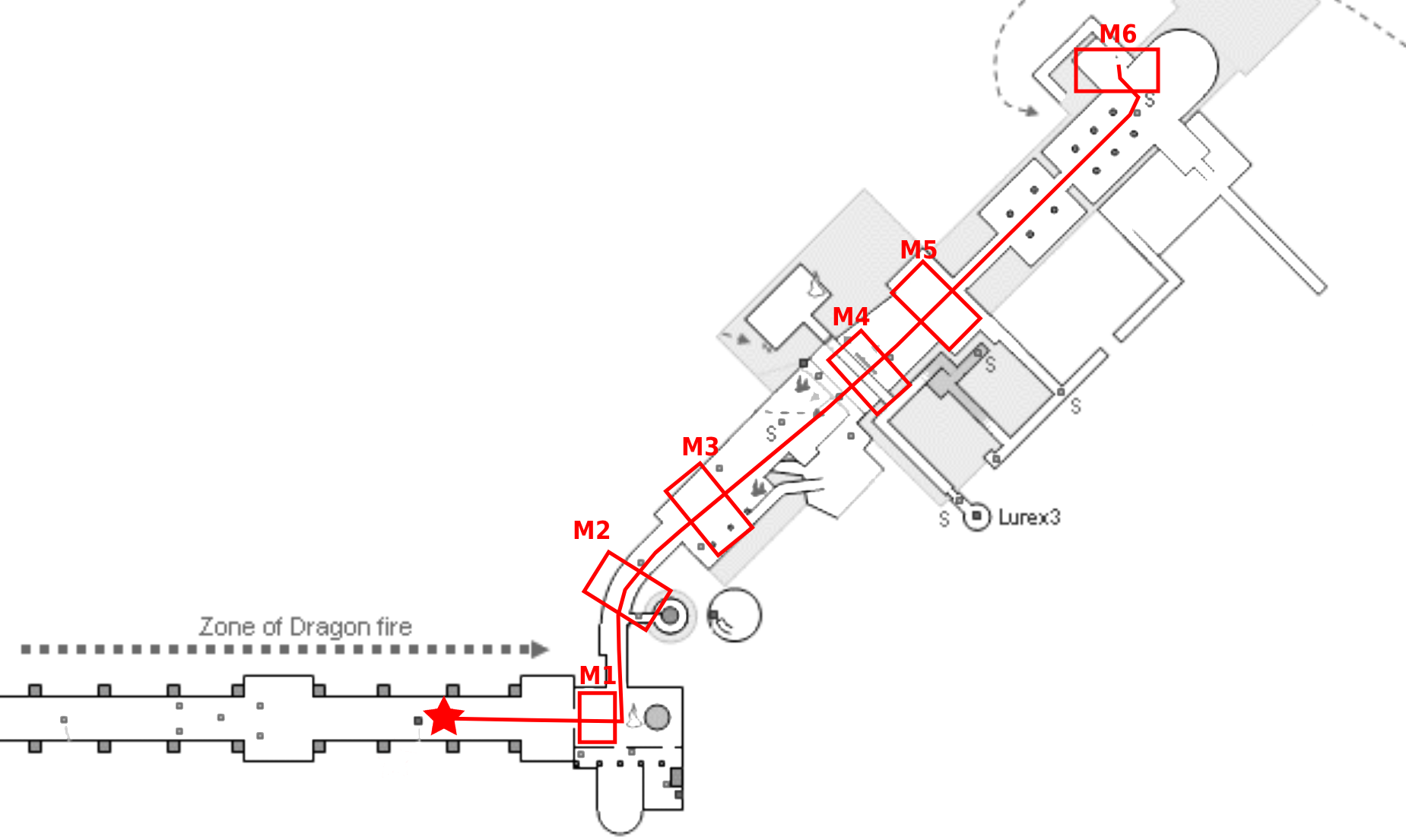}
  \caption{Dark Souls~I Undead Parish (DS1-UP) route used in our experiments. The magenta polyline shows the evaluation path from the dragon bridge toward the Gargoyle church; magenta rectangles mark the six visual milestones
  $M_1$-$M_6$.}
  \label{fig:ds1-up-route}
\end{figure}

\section{Additional Route Maps and Milestones}
\label{sec:route-maps}

For space reasons, the main text only visualizes the Grand Archives and Black Myth routes. Figures here provide analogous top-down maps for the remaining routes in our pilot study, with the designer-specified paths and milestones $M_1$–$M_6$ overlaid for reference.

\end{document}